\documentclass{article}

\PassOptionsToPackage{numbers, compress}{natbib}

    \usepackage[preprint]{neurips_2023}

\usepackage[utf8]{inputenc} 
\usepackage[T1]{fontenc}    
\usepackage[pagebackref=true,breaklinks=true,letterpaper=true,colorlinks,bookmarks=false]{hyperref}   
\usepackage{url}            
\usepackage{booktabs}       
\usepackage{amsfonts}       
\usepackage{nicefrac}       
\usepackage{microtype}      
\usepackage[dvipsnames]{xcolor}
\usepackage[pdftex]{graphicx}
\usepackage{colortbl}
\usepackage{multirow}
\usepackage{amssymb}
\usepackage{amsmath}
\usepackage{wrapfig}
\usepackage{float}

\definecolor{mygreen}{rgb}{0.44, 0.68, 0.28}
\definecolor{myblue}{rgb}{0.36, 0.61, 0.84}
\definecolor{myorange}{rgb}{0.93, 0.49, 0.19}

\def\onedot{. }
 
\def\ie{\emph{i.e}\onedot}

\def\etal{\emph{et al}\onedot}

\usepackage[T1]{fontenc}

\usepackage{mathabx}
\def\CircleArrowright{\ensuremath{\rotatebox[origin=c]{310}{$\circlearrowright$}}}
\newcommand{\rvlnbert}{VLN$\protect\CircleArrowright$BERT}

\title{NavGPT: Explicit Reasoning in Vision-and-Language Navigation with Large Language Models}

\author{%
  Gengze Zhou$^1$\quad
  Yicong Hong$^2$\quad
  Qi Wu$^1$ \\
  $^1$The University of Adelaide\quad
  $^2$The Australian National University \\
  \texttt{\{gengze.zhou, qi.wu01\}@adelaide.edu.au}\quad
  \texttt{yicong.hong@anu.edu.au}\\
{\small \url{https://github.com/GengzeZhou/NavGPT}}
}

\begin{document}

\maketitle

\begin{abstract}
 Trained with an unprecedented scale of data, large language models (LLMs) like ChatGPT and GPT-4 exhibit the emergence of significant reasoning abilities from model scaling. Such a trend underscored the potential of training LLMs with unlimited language data, advancing the development of a universal embodied agent. 
 In this work, we introduce the NavGPT, a purely LLM-based instruction-following navigation agent, to reveal the reasoning capability of GPT models in complex embodied scenes by performing zero-shot sequential action prediction for vision-and-language navigation (VLN).
 At each step, NavGPT takes the textual descriptions of visual observations, navigation history, and future explorable directions as inputs to reason the agent's current status, and makes the decision to approach the target.
 Through comprehensive experiments, we demonstrate NavGPT can explicitly perform high-level planning for navigation, including decomposing instruction into sub-goal, integrating commonsense knowledge relevant to navigation task resolution, identifying landmarks from observed scenes, tracking navigation progress, and adapting to exceptions with plan adjustment. 
 Furthermore, we show that LLMs is capable of generating high-quality navigational instructions from observations and actions along a path, as well as drawing accurate top-down metric trajectory given the agent's navigation history. Despite the performance of using NavGPT to zero-shot R2R tasks still falling short of trained models, we suggest adapting multi-modality inputs for LLMs to use as visual navigation agents and applying the explicit reasoning of LLMs to benefit learning-based models.
\end{abstract}

\section{Introduction}

Amid the remarkable advances in large language model (LLM) training \cite{touvron2023llama, brown2020language, chowdhery2022palm, zhang2022opt, wei2021finetuned, vicuna2023, bubeck2023sparks, openai2023gpt4}, we note a shift towards integrating LLMs into embodied robotics tasks such as SayCan~\cite{saycan2022arxiv} and PaLM-E~\cite{driess2023palm}. This trend stems from two primary considerations: the scale of training data and the scale of models. First, the development of techniques for processing textual information provides an abundant source of natural language training data for learning interdisciplinary and generalizable knowledge. Furthermore, by accessing unlimited language data, significant emergent abilities~\cite{wei2022emergent} are observed when scaling up the model, resulting in a remarkable enhancement in the reasoning capabilities when solving problems across wide domains. Consequently, training an LLM with unlimited language data is seen as a viable pathway toward realizing a universal embodied agent. 

\begin{figure}
	\centering
\includegraphics[width=.99\linewidth]{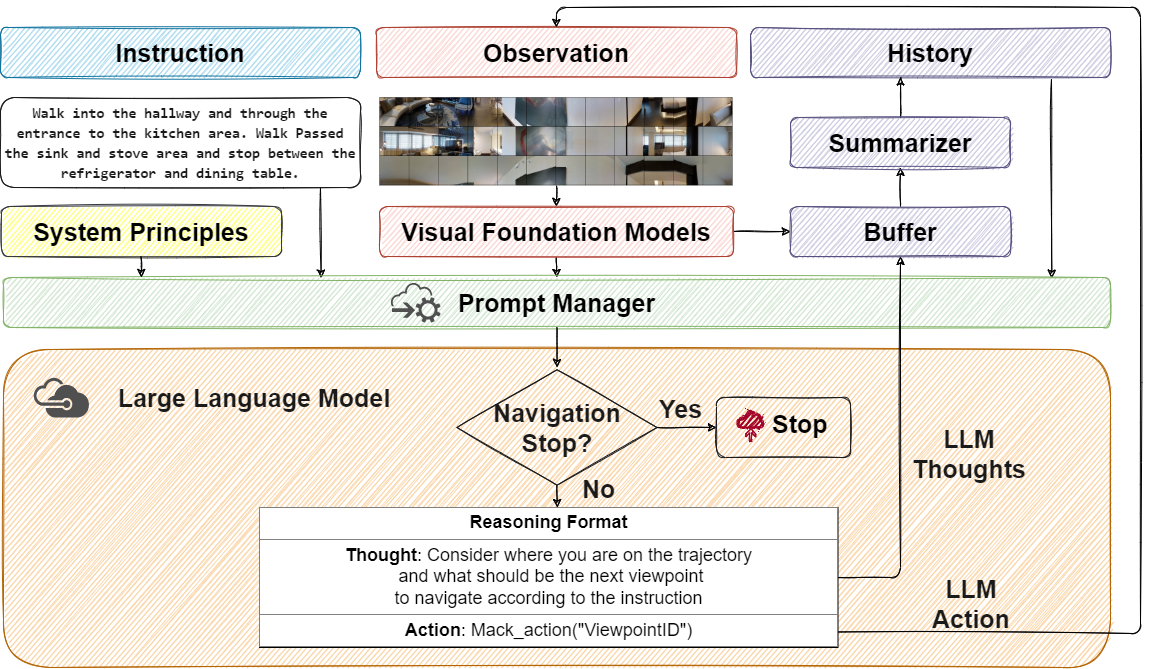}
	\caption{\small The architecture of NavGPT. NavGPT synergizes reasoning and actions in LLMs to perform zero-shot Vision-and-Language Navigation following navigation system principles. It interactives with different visual foundation models to adapt multi-modality inputs, handle the length of history with a history buffer and a GPT-3.5 summarizer, and aggregate various sources of information through a prompt manager. NavGPT parse the generated results from LLMs (LLM \textit{Thoughts} and LLM \textit{Action}) to move to the next viewpoint.}
	\label{fig:model_architecture}
\vspace{-1em}
\end{figure}

This insight has spurred the integration of LLMs into vision-and-language navigation (VLN)~\cite{anderson2018vision}, an exploratory task toward achieving real-world instruction-following embodied agents. The latest research attempt to leverage GPT models~\cite{openai2023gpt4, brown2020language} to benefit navigation. For example, using LLMs as a parser for diverse language input~\cite{shah2023lm} — extracting landmarks from instruction to support visual matching and planning, or leveraging LLMs' commonsense reasoning abilities ~\cite{zhou2023esc, dorbala2023can} to incorporate prior knowledge of inter-object correlations to extend agents' perception and facilitate the decision making. However, we notice that the reasoning ability of LLMs in navigation is still under-explored, \ie, can LLMs understand the interactive world, the actions, and consequences in text form, and use all the information to solve a navigation task?

In light of this, we introduce NavGPT, a fully automatic LLM-based system designed for language-guided visual navigation, with the capability to handle multi-modality inputs, unconstrained language guidance, interaction with an open-world environment, and progress tracking with navigation history. NavGPT perceives the visual world by reading descriptions of observations generated by visual foundation models (VFMs), and synergizing \textit{Thoughts} (reasoning) and \textit{Actions} (decision making) in an explicit text form. To an extreme extent, we use NavGPT to perform zero-shot VLN\footnote{Our NavGPT is solely powered by off-the-shelf LLMs, without any learnable module or any prior experience in solving interactive navigation. Hence, all navigation tasks defined in this paper are novel to NavGPT.} to clearly reveal the reasoning process of LLMs during navigation.

Through comprehensive experiments, we found that LLMs possess the capability to execute complex navigational planning. This includes the deconstruction of instructions into distinct sub-goals, assimilation of commonsense knowledge pertinent to navigational tasks, identification of landmarks within the context of observed environments, continuous monitoring of navigational progression, and responding to anomalies by modifying their initial plan. The aforementioned phenomenon reflects an astonishing reasoning ability in understanding and solving navigation problems. Furthermore, we show that LLMs have the ability to draw navigation trajectories in a metric map and regenerate navigation instruction based on navigation history, revealing the historical and spatial awareness of LLMs for navigation tasks. However, there remains a significant gap between the zero-shot performance of current open-sourced LLMs in VLN compared to the fine-tuned models, where the bottleneck of NavGPT lies in the information loss while translating visual signals into natural language and summarizing observations into history. As a result, we suggest the future direction of building general VLN agents to be LLMs with multi-modality inputs or a navigation system making use of high-level navigation planning, historical and spatial awareness from LLMs.

Our contributions can be summarized as follow: (1) We introduce a novel instruction-following LLMs agent for visual navigation with a supportive system to interact with the environment and track navigation history. (2) We investigate the capabilities and limitations of current LLMs' reasoning for making navigation decisions. (3) We reveal the capability of LLMs in high-level planning for navigation, by observing the thoughts of LLMs, making the planning process of navigation agents accessible and explainable.

\section{Related Work}

\paragraph{Vision-and-Language Navigation}
Language-driven vision navigation is demanded by widely applicable embodied navigation agents. Previous study shows the essentials of modules to achieve such a goal ~\cite{anderson2018vision,qi2020reverie,krantz2020beyond,ku2020room,he2021landmark,gu2022vision,wang2018look,Zhu2022diagnosing, hong2020language, hong2022bridging}, whereas a large number of research reveal the crucial effect of training strategies~\cite{wang2019reinforced,tan2019learning}.
Importantly, the main problem lies in VLN is the generalizability of agents in unseen environments. 
Data augmentation~\cite{liu2021vision,wang2022less,li2022envedit,tan2019learning,parvaneh2020counterfactual,fu2020counterfactual,wang2022counterfactual}, 
memory mechanism~\cite{chen2021history,wang2021structured,pashevich2021episodic}, 
pre-training~\cite{majumdar2020improving,hao2020towards,guhur2021airbert,wu2022cross,qi2021road} 
have been adopted to alleviate data scarcity. However, those augmentations and pre-training are limited to the sampled data from a fixed number of scenes, which is not enough to reflect a realistic application scene where objects could be out of the domains and language instructions are more diverse. In our work, we utilize the reasoning and knowledge storage of LLMs and perform VLN in a zero-shot manner as an initial attempt to reveal the potential usage of LLMs for VLN in the wild.
A number of studies ~\cite{chen2021topological,deng2020evolving,chen2022think,wang2021structured} have presented compelling methodologies that underscore the significance of topological maps in facilitating long-term planning, specifically in the aspect of backtracking to prior locations. In addition, Dorbala~\etal~\cite{dorbala2022clip} use CLIP~\cite{radford2021learning} to perform zero-shot VLN by chunking instructions into keyphrases and completely rely on the text-image matching capability from CLIP to navigate.
However, the planning and decision making processes of the agents above are implicit and not accessible. On the contrary, benefiting from the intrinsic of LLMs, we are able to access the reasoning process of agents, making it explainable and controllable.

\paragraph{Large language models.}
With the massive success in large-scale language model training \cite{touvron2023llama, brown2020language, chowdhery2022palm, zhang2022opt, wei2021finetuned, vicuna2023}, a new cohort of Large Language Models (LLMs) has shown evolutionary progress toward achieving Artificial General Intelligence (AGI)~\cite{bubeck2023sparks, openai2023gpt4}. This burgeoning class of LLMs, underpinned by increasingly sophisticated architectures and training methodologies~\cite{vicuna2023, scao2022language}, has the potential to revolutionize various domains by offering unprecedented capabilities in natural language understanding and generation. The main concern for LLMs is that their knowledge is limited and confined after training is finished. The latest works study how to utilize LLMs interacting with tools to expand their knowledge as a plugin, including extending LLM to process multimodality content~\cite{wu2023visual, shen2023hugginggpt}, teaching LLMs to access the internet with correct API calls~\cite{schick2023toolformer}, and expanding their knowledge with local databases to accomplish QA tasks~\cite{peng2023check}. Another stream of works studies how to prompt LLMs in a hierarchical system to facilitate the alignment of reasoning and corresponding actions~\cite{yao2022react, karpas2022mrkl} beyond the Chain of Thought (CoT)~\cite{wei2022chain}. These works set up the preliminaries for building an embodied agent directly using LLMs.

\paragraph{LLMs in Robotics Navigation.}
The employment of Large Language Models (LLMs) in the field of robotics remains in the primary stage ~\cite{vemprala2023chatgpt, bubeck2023sparks}. A handful of contemporary studies, however, have begun to explore the utilization of generative models for navigation. Shah \etal ~\cite{shah2023lm} employs GPT-3~\cite{brown2020language} in an attempt to identify "landmarks" or subgoals, while Huang \etal ~\cite{huang2022visual} concentrates its efforts on the application of an LLM for the generation of code. Zhou \etal ~\cite{zhou2023esc} use LLM to extract the commonsense knowledge of the relations between targets and objects in observations to perform zero-shot object navigation (ZSON)~\cite{gadre2022cow, majumdar2022zson}. Despite these recent advancements, our study diverges in its concentration on converting visual scene semantics into input prompts for the LLM, directly performing VLN based on the commonsense knowledge and reasoning ability of LLMs. The work closest to ours is LGX ~\cite{dorbala2023can}, but they are doing object navigation where agents are not required to follow the instruction and in their method, they use the GLIP~\cite{li2022grounded} model to decide the stop probability and did not consider memorization of navigation history, action, and reasoning between LLM.

\section{Method}
\noindent\textbf{VLN Problem Formulation.} We formulate the VLN problem as follows: Given a natural language instruction $\mathcal{W}$, composed of a series of words $\{w_1, w_2, w_3, \dots, w_n\}$, at every step $s_t$, the agent interprets the current location via the simulator to obtain an observation $\mathcal{O}$. This observation comprises $N$ alternative viewpoints, representing the egocentric perspectives of agents in varying orientations.

Each unique view observation is denoted as $o_i (i\leq N)$, with its associated angle direction represented as $a_i (i \leq N)$. The observation can thus be defined as $\mathcal{O}_t \triangleq [\langle o_1, a_1 \rangle, \langle o_2, a_2 \rangle, \dots, \langle o_N, a_N \rangle]$. Throughout the navigation process, the agents' action space is confined to the navigation graph $G$. The agent must select from the $M = |C_{t+1}|$ navigable viewpoints, where $C_{t+1}$ indicates the set of candidate viewpoints, by aligning the observation $\mathcal{O}_t^C \triangleq [\langle o^C_1, a^C_1 \rangle, \langle o^C_2, a^C_2 \rangle, \dots, \langle o^C_{M}, a^C_{M} \rangle]$ with the oracle $\mathcal{W}$. The agent prognosticates the subsequent action by selecting the relative angle $a^C_i$ from $\mathcal{O}_t^C$, then enacts this action through interaction with the simulator to transition from the current state $s_t = \langle v_t, \theta_t, \phi_t \rangle$ to $s_{t+1} = \langle v_{t+1}, \theta_{t+1}, \phi_{t+1} \rangle$, where $v$, $\theta$ and $\phi$ denotes the current viewpoint location, the current heading and elevation angle of the agent respectively. The agent also maintains a record of the state history $h_t$ and adjusts the conditional transition probability between states $\mathcal{S}_t = T(s_{t+1}|a^C_i, s_t, h_t)$, where function $T$ denotes the conditional transition probability distribution.

In summary, the policy $\pi$ parametrized by ${\Theta}$ that the agent is required to learn is based on the oracle $\mathcal{W}$ and the current observation $\mathcal{O}_t^C$, which is $\pi(a_t|\mathcal{W}, \mathcal{O}_t, \mathcal{O}_t^C, \mathcal{S}_t ; {\Theta})$. In this study, NavGPT conducts the VLN task in a zero-shot manner, where the ${\Theta}$ is not learned from the VLN datasets, but from the language corpus that the LLMs are trained on.

\subsection{NavGPT}
NavGPT is a system that interacts with environments, language guidance, and navigation history to perform action prediction. Let $\mathcal{H}_{<t+1} \triangleq [\langle \mathcal{O}_1, \mathcal{R}_1, \mathcal{A}_1 \rangle, \langle \mathcal{O}_2, \mathcal{R}_2, \mathcal{A}_2 \rangle, \dots, \langle \mathcal{O}_t, \mathcal{R}_t, \mathcal{A}_t \rangle]$ be the navigation history of observation $\mathcal{O}$, LLM reasoning $\mathcal{R}$ and action $\mathcal{A}$ triplets for the previous $t$ steps. To obtain the navigation decision $\mathcal{A}_{t+1}$, NavGPT needs to synergize the visual perception from VFMs $\mathcal{F}$, language instruction $\mathcal{W}$, history $\mathcal{H}$ and navigation system principle $\mathcal{P}$ with the help of prompt manager $\mathcal{M}$, define as follow:
\begin{equation}
    \langle \mathcal{R}_{t+1}, \mathcal{A}_{t+1} \rangle = LLM(\mathcal{M}(\mathcal{P}), \mathcal{M}(\mathcal{W}), \mathcal{M}(\mathcal{F}(\mathcal{O}_t)), \mathcal{M}(\mathcal{H}_{<t+1}))
\end{equation}

\noindent\textbf{Navigation System Principle $\mathcal{P}$}.
The Navigation System Principle formulates the behavior of LLM as a VLN agent. It clearly defines the VLN task and the basic reasoning format and rules for NavGPT at each navigation step. For example, NavGPT should move among the static viewpoints (positions) of a pre-defined graph of the environment by identifying the unique viewpoint ID. NavGPT should not fabricate nonexistent IDs. Details are discussed in section \ref{sec:manager}.

\noindent\textbf{Visual Foundation Models $\mathcal{F}$}.
NavGPT as an LLM agent requires visual perception and expression ability from VFMs to translate the current environment's visual observation into natural language description. The VFMs here play the role of translator, to translate visual observations using their own language, \textit{e.g.} natural language, objects' bounding boxes, and objects' depth. Through the process of prompt management, the visual perception results will be reformated and translated into pure natural language for LLMs to understand, discussed in section \ref{sec:VFM}.

\noindent\textbf{Navigation History $\mathcal{H}_{<t+1}$}.
The navigation history is essential for NavGPT to evaluate the progress of the completion of the instruction, to update the current state, and make the following decisions. The history is composed of summarized descriptions of previous observations $\mathcal{O}_{<t+1}$ and actions $\mathcal{A}_{<t+1}$, along with the reasoning thoughts $\mathcal{R}_{<t+1}$ from LLM, discussed in section \ref{sec:reasoning}.

\noindent\textbf{Prompt Manager $\mathcal{M}$}.
The key to using LLM as a VLN agent is to convert all the above content into a natural language that LLM can understand. This process is done by the prompt manager, which collects the results from different components and parses them into a single prompt for LLM to make navigation decisions, discussed in section \ref{sec:manager}.

\subsection{Visual Perceptron for NavGPT}\label{sec:VFM}
\begin{figure}
	\centering
\includegraphics[width=.99\linewidth]{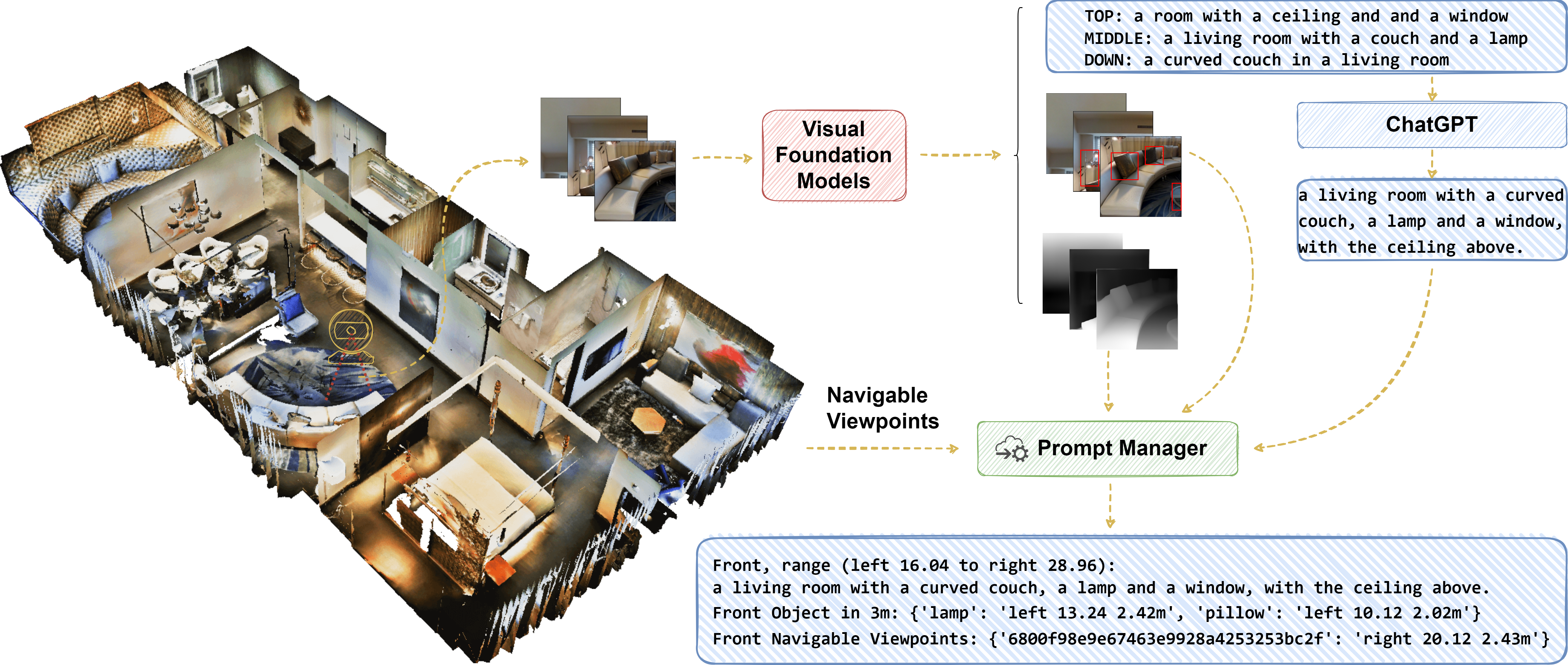}
	\caption{\small The process of forming natural language description from visual input. We used 8 directions to represent a viewpoint and show the process of forming the descriptions for one of the directions.}
	\label{fig:obs}
\vspace{-1em}
\end{figure}

In this section, we introduce the visual perception process of NavGPT. We take visual signals as a foreign language and handle the visual input using different visual foundation models to translate them into natural language, shown in figure \ref{fig:obs}.

For an agent standing at any viewpoint in the environment, the observation is composed of egocentric views from different orientations. The number of total views is defined by the field of view of each view image and the relative angle of each view. In our work, we set the field of view of each view as $45^\circ$, and turn the heading angle $\theta$ $45^\circ$ per view from $0^\circ$ to $360^\circ$, 8 directions in total. Besides, we turn the elevation angle $\phi$ $30^\circ$ per view from $30^\circ$ above the horizontal level to $30^\circ$ below, 3 levels in total. As a result, we obtain $3 \times 8 = 24$ egocentric views for each viewpoint.

To translate visual observation into natural language, we first utilize the BLIP-2~\cite{li2023blip} model as the translator. With the strong text generation capability of LLMs, BLIP-2 can achieve stunning zero-shot image-to-text generation quality. By carefully setting the granularity of visual observation (field of views and the total view number in each observation), we prompt BILP-2 to generate a decent language description of each view with a detailed depiction of the shapes and color of objects and the scenes they are in while avoiding useless caption of views from a smaller FoV, from which partial observation is available and it is hard to recognize even for humans. See appendix for details.

Notice that for the heading direction, the rotation interval is equal to the field of view, therefore there is no overlapping between each orientation. For the elevations, there is a $15^{\circ}$'s overlapping between the top, middle, and down views. In NavGPT we mainly focus on the heading angle of agents during navigation, therefore, we prompt GPT-3.5 to summarize the scenes from the top, middle, and down views for each orientation into a sentence of description.

Besides natural language descriptions of the scene from BLIP-2, we also excavate the lower-level feature extracted by other vision models. These vision models serve as auxiliary translators, translating visual input into their own "language" like the class of objects and corresponding bounding boxes. The detection results will be aggregated by the prompt manager into prompts for LLMs. In this work, we utilize Fast-RCNN~\cite{girshick2015fast} to extract the bounding boxes of objects in each egocentric view. After locating the objects, we calculate the relative heading angle for each object and the agent. We also extract the depth information of the center pixel of the object provided by the Matterport3D simulator~\cite{anderson2018vision}. With the depth, objects' relative orientation, and class, we filter the detection results by leaving the object within 3 meters from the current viewpoint. The results from VFMs will be processed by the prompt manager into observation for the current viewpoint in natural language.

\subsection{Synergizing reasoning and actions in LLMs}\label{sec:reasoning}
In the VLN task, the agent needs to learn the policy $\pi(a_t|\mathcal{W}, \mathcal{O}_t, \mathcal{O}_t^C, \mathcal{S}_t; {\Theta})$, which is difficult because the implicit connection between actions and observations and demain intensive computation. In order to explicitly access and enhance the agent's comprehension of the current state during navigation, we follow the ReAct paper \cite{yao2022react} to expand the agent's action space to $\tilde{\mathcal{A}} = \mathcal{A} \cup \mathcal{R}$, where $\mathcal{R} \in \mathcal{L}$ is in the entire language space $\mathcal{L}$, denoting the thought or reasoning trace of the agent.

The reasoning traces $\mathcal{R}$ of the agent will not trigger any interaction with the external environment, therefore no observation will be returned when the agent is outputting the reasoning during each navigation step. We synergize the NavGPT's actions and thoughts by prompting it to make navigation decisions after outputting the reasoning trace at each step. Introducing the reasoning traces aims to bootstrap the LLMs in two aspects: 

Firstly, prompting the LLMs to think before choosing an action, enables LLMs to perform complex reasoning in planning and creating strategies to follow the instructions under the new observations. For example, as shown in figure \ref{fig:qualitative}, NavGPT can generate a long-term navigation plan by analyzing the current observation and the instruction, performing higher-level planning such as decomposing instruction and planning to reach the sub-goal, which is never seen explicitly in previous works. 

Secondly, including reasoning traces $\mathcal{R}$ in the navigation history $\mathcal{H}_{<t}$ enhances the problem-solving ability of NavGPT. By injecting reasoning traces into navigation history, NavGPT inherits from the previous reasoning traces, to reach a sub-goal with high-level planning consistently through steps, and can track the navigation progress with exception-handling abilities like adjusting the plan.

\subsection{NavGPT prompt manager}\label{sec:manager}
With the Navigation System Principle $\mathcal{P}$, translated results from VFMs, and the History of Navigation $\mathcal{H}_{<t}$, the prompt manager parses and reformates them into prompts for LLMs. Details of the prompt are presented in the appendix.

Specifically, for Navigation System Principle $\mathcal{P}$, NavGPT prompt manager will create a prompt to convey LLMs with the rules, declaring the VLN task definition, defining the simulation environment for NavGPT, and restricting LLMs' behavior in the given reasoning format.

For perception results from VFMs $\mathcal{F}$, the prompt manager gathers the results from each direction and orders the language description by taking the current orientation of NavGPT as the front, shown in figure \ref{fig:obs}, arranging the description from 8 directions into prompt by concatenating them clockwise.

For navigation history $\mathcal{H}_{<t+1}$, the observation, reasoning, and actions triples $\langle \mathcal{O}_i, \mathcal{R}_i, \mathcal{A}_i \rangle$  are stored in a history buffer, shown in figure \ref{fig:model_architecture}. Directly extracting all triples in the buffer will create too long a prompt for LLMs to accept. To handle the length of history, the prompt manager utilizes GPT-3.5 to summarize the observations from viewpoints in the trajectory, inserting the summarized observations into the observation, reasoning, and actions triples in the prompt.

\section{Experiment}

\noindent\textbf{Implementation Details.}
We evaluate NavGPT based on GPT-4~\cite{openai2023gpt4} and GPT-3.5 on the R2R-VLN dataset~\cite{anderson2018vision}. The R2R dataset is composed of 7189 trajectories, each corresponding to three fine-grained instructions. The dataset is separated into the train, val seen, val unseen, and test unseen splits, with 61, 56, 11, and 18 indoor scenes, respectively. We apply the 783 trajectories in the 11 val unseen environments in all our experiments and for comparison to previous supervised approaches.
We utilize BLIP-2 ViT-G $\text{FlanT5}_\text{XL}$~\cite{li2023blip} as images translator and Faster-RCNN~\cite{girshick2015fast} as object detector. The depth information of objects is extracted from the Mattport3D simulator~\cite{anderson2018vision} by taking the depth of the center pixel in the bounding box.

\noindent\textbf{Evaluation Metrics.}
The evaluation of NavGPT utilizes standardized metrics from the R2R dataset. These include Trajectory Length (TL), denoting the average distance traveled by the agent; Navigation Error (NE), representing the mean distance from the agent's final location to the destination; Success Rate (SR), indicating the proportion of navigation episodes where the agent successfully reaches the target location within a 3-meter margin of error; Oracle Success Rate (OSR), the success rate of agent stopped at the closest point to the goal on its trajectory; and Success Rate weighted by the normalized inverse of Path Length (SPL), which is a more nuanced measure that balances navigation precision and efficiency by adjusting the success rate based on the ratio of the optimal path length to the agent's predicted path length.

\subsection{Qualitive Results}

We elaborately study the qualitative results of the reason trace from NavGPT. We reveal the potential high-level planning capability of GPT-4 under embodied navigation tasks.

\begin{figure}[t]
	\centering
	\includegraphics[width=.99\linewidth]{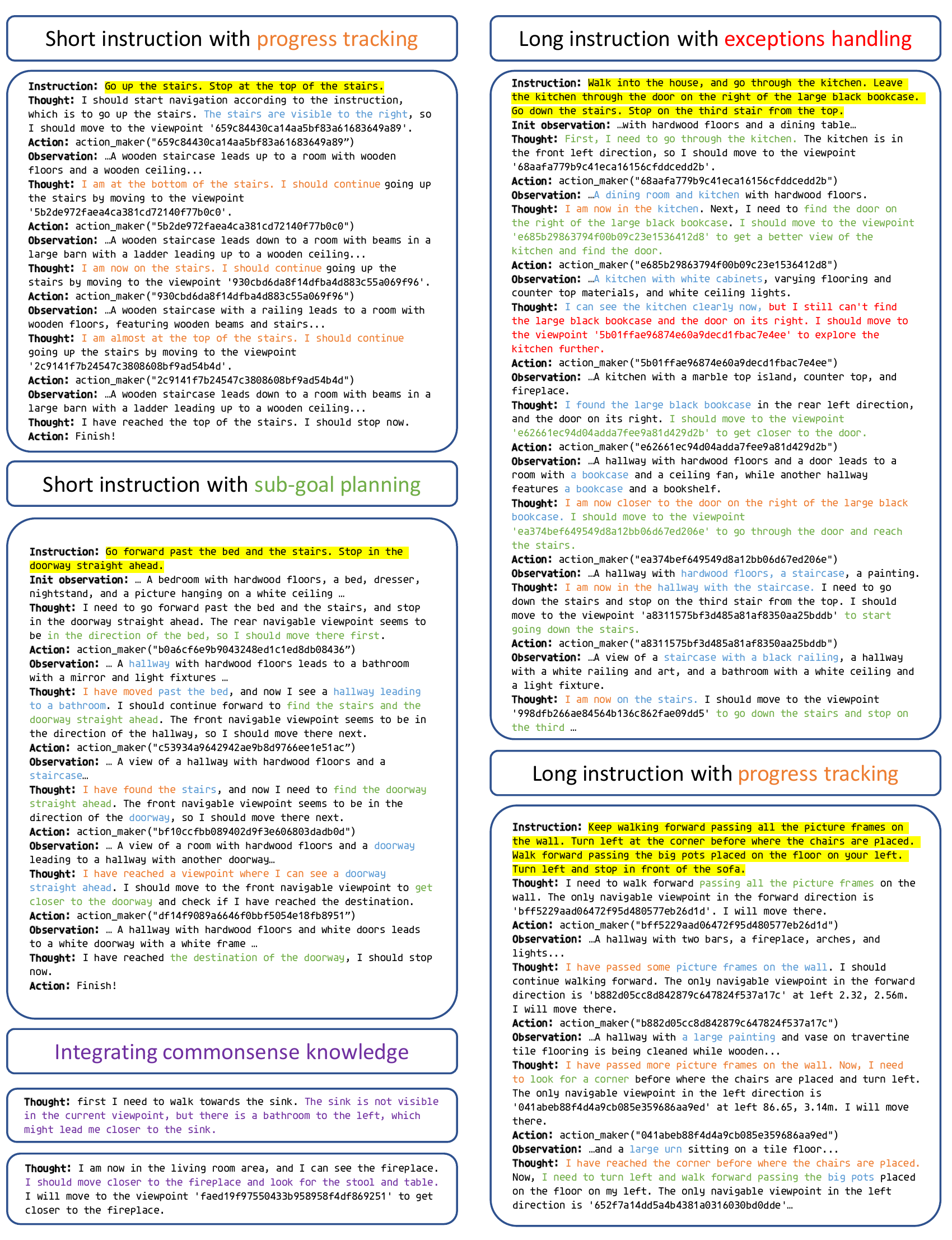}
	\caption{\small The qualitative of NavGPT. NavGPT can explicitly perform high-level planning for sequential action prediction, including \textcolor{mygreen}{decomposing instruction into sub-goal}, \textcolor{Orchid}{integrating commonsense knowledge}, \textcolor{myblue}{identifying landmarks from observed scenes}, \textcolor{myorange}{tracking navigation progress}, \textcolor{red}{exceptions handling with plan adjustment}.}
	\label{fig:qualitative}
	\vspace{-1.em}
\end{figure}

\paragraph{Reasoning capability of GPT-4 for Language-guide Navigation}
As shown in figure \ref{fig:qualitative}, with GPT-4, NavGPT can perform various types of reasoning and high-level planning during navigation. For short instructions, NavGPT can \textcolor{myorange}{track the navigation progress} through steps to accomplish a single action described in the instructions, similar to the self-monitoring VLN agents~\cite{ma2019self, zhu2020vision}. For long instructions, NavGPT can \textcolor{mygreen}{break it down with sub-goals}, similar to previous works on fine-graining R2R data~\cite{hong2020sub, he2021landmark, zhu2020babywalk}, and plan to reach the destination by effectively \textcolor{myblue}{identifying landmarks from observations}, similar to works on utilizing objects information to perform cross-modality matching in VLN \cite{gao2021room, qi2020object, qi2021road}. When navigating to a viewpoint with \textcolor{red}{unexpected observation}, NavGPT can \textcolor{red}{plan to explore} the environment and use \textcolor{Orchid}{commonsense knowledge to assist decision-making}, similar to VLN methods incorporate external knowledge\cite{li2022incorporating, gao2021room, VLN_LAD_2023}.

\paragraph{History and spatial relative relation awareness for LLMs during navigation}
We examined NavGPT's awareness of historical and spatial relations by employing GPT-4 to delineate the trajectory in navigational history and to construct a map of visited viewpoints utilizing pyplot. The process involved extracting exclusively the actions $\mathcal{A}_{t+1}$, observations $\mathcal{O}_{t+1}$, and the entire navigation history $\mathcal{H}_{t+1}$. The specifics of the prompt are presented in the appendix.

\begin{figure}[t]
	\centering
	\includegraphics[width=.99\linewidth]{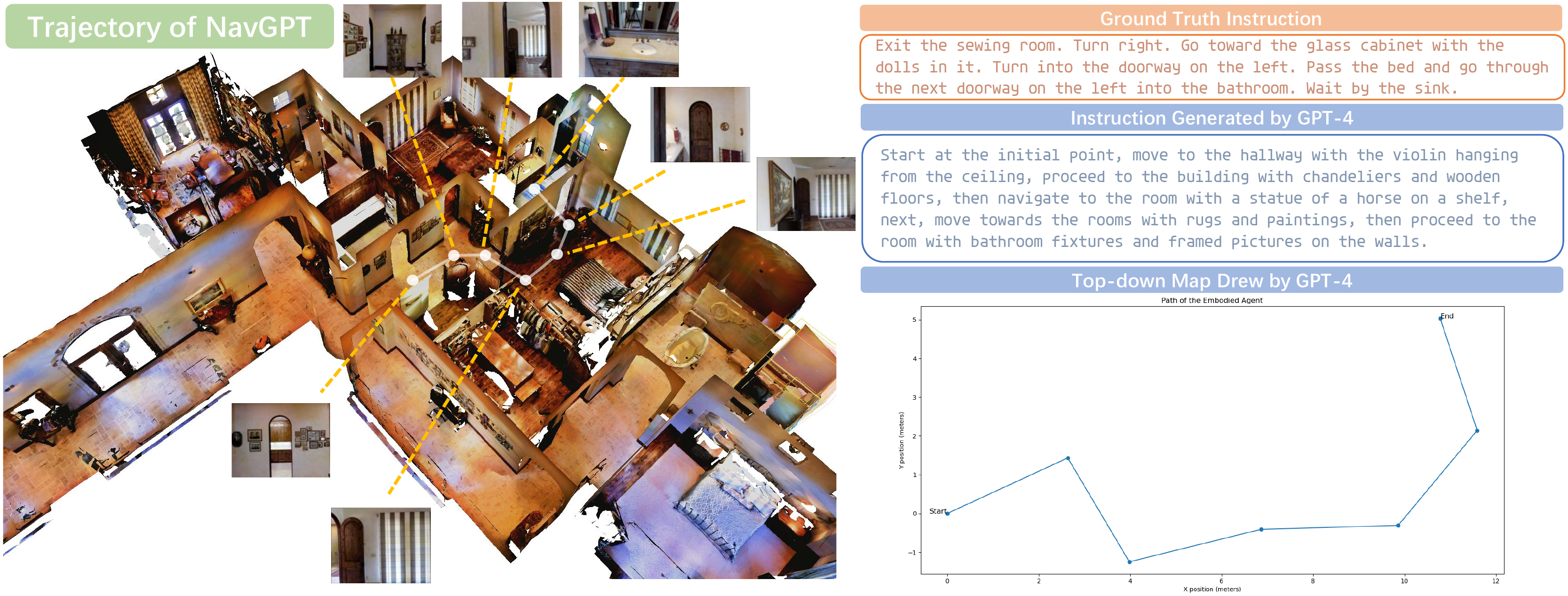}
	\caption{\small We evaluate GPT-4 on a case where NavGPT successfully follows the ground truth path, using only the historical actions $\mathcal{A}_{<t+1}$ and observations $\mathcal{O}_{<t+1}$ to generate an instruction (without reasoning trace $\mathcal{R}_{<t+1}$ to avoid information leaking), and using the entire navigation history $\mathcal{H}_{<t+1}$ to draw a top-down trajectory.}
	\label{fig:qualitative2}
	\vspace{-1.em}
\end{figure}

As shown in figure \ref{fig:qualitative2}, we observed that GPT-4 could effectively extract landmarks from the redundant observation descriptions and generate navigation history descriptions with actions. This could be a potential way of generating new trajectory instructions for VLN. Besides, the result shows GPT-4 can comprehensively understand the history of navigation, and thus can perform the essential progress tracking during navigation. Moreover, shown in figure \ref{fig:qualitative2}, GPT-4 can successfully catch the relative position relations between viewpoints and draw a top-down view of the trajectory for visited viewpoints. By providing language descriptions of actions taken by the agents, including the turning angle and relative distances between viewpoints, GPT-4 shows a stunning awareness of spatial relations. Such impressive reasoning ability support NavGPT to perform high-level planning shown in figure \ref{fig:qualitative}, underlines the significant potential LLMs hold for embodied navigation tasks.

\subsection{Comparison with Supervised Methods}

We compare the results of using NavGPT with GPT-4 to zero-shot the sequential navigation tasks with previous models trained on the R2R dataset. As shown in table \ref{tab:main}, 
a significant discrepancy can be discerned. We suggest the limitations inhibiting the performance of LLMs in solving VLN can be primarily attributed to two factors: the precision of language-based depiction of visual scenes and the tracking capabilities regarding objects.

\definecolor{Gray}{gray}{0.9}
\begin{table}[ht]
\small
\centering
\caption{Comparison with previous methods on R2R validation unseen split.}
\label{tab:main}
\resizebox{0.75\linewidth}{!}
{
\begin{tabular}{llccc>{\columncolor{Gray}}c>{\columncolor{Gray}}c}
\toprule
\multicolumn{1}{c}{Training Schema} & \multicolumn{1}{c}{Method} &
\multicolumn{1}{c}{TL} & \multicolumn{1}{c}{NE$\downarrow$} & \multicolumn{1}{c}{OSR$\uparrow$} & \multicolumn{1}{c}{SR$\uparrow$} & \multicolumn{1}{c}{SPL$\uparrow$} \\
\midrule
\multirow{3}{*}{Train Only} 
& Seq2Seq~\cite{anderson2018vision}       & 8.39  & 7.81 & 28 & 21 & - \\
& Speaker Follower~\cite{fried2018speaker} & - & 6.62 & 45 & 35 & - \\
& EnvDrop~\cite{tan2019learning} & 10.70 & 5.22 & - & 52 & 48 \\
\midrule
\multirow{4}{*}{Pretrain + Finetune} 
& PREVALENT~\cite{hao2020towards} & 10.19 & 4.71 & - & 58 & 53 \\
& \rvlnbert~\cite{hong2021vln}   & 12.01 & 3.93 & 69 & 63 & 57 \\
& HAMT~\cite{chen2021history}      & 11.46 & 2.29 & 73 & 66 & 61 \\
& DuET~\cite{chen2022think}      & 13.94 & 3.31 & 81 & 72 & 60 \\
\midrule
\multirow{2}{*}{No Train} 
& DuET (Init. LXMERT~\cite{tan2019lxmert}) & 22.03 & 9.74 & 7 & 1 & 0 \\
& NavGPT (Ours) & 11.45 & 6.46 & 42 & 34 & 29 \\
\bottomrule
\end{tabular}
}
\end{table}

NavGPT's functionality is heavily reliant on the quality of captions generated from VFMs. If the target object delineated in the instruction is absent in the observation description, NavGPT is compelled to explore the environment. The ideal circumstance entails all target objects being visible pursuant to the instruction. However, the inherent granularity of language description inevitably incurs a loss of information. Moreover, NavGPT must manage the length of the navigation history to prevent excessively verbose descriptions as the steps accrue. To this end, a summarizer is implemented, albeit at the cost of further information loss. This diminishes NavGPT's tracking ability, impeding the formation of seamless perceptions of the entire environment as the trajectory lengthens.

\subsection{Effect of Visual Components}

We perform additional experiments to investigate the effectiveness of visual components in NavGPT, we construct a baseline with GPT-3.5 for its easier access and budget-friendly costs. To evaluate the zero-shot ability in various environments, we construct a new validation split sampling both from the original training set and the validation unseen set. The scenes from the training and validation unseen set are 61 and 11 respectively, 72 scenes in total. We randomly picked 1 trajectory from the 72 environment, each is associated with 3 instructions. In total, we sample 216 samples to conduct the ablation study.

\vspace{-0.5em}
\paragraph{Effect of granularity in visual observation descriptions.}

\definecolor{Gray}{gray}{0.9}
\begin{wraptable}{r}{0.55\linewidth}
\centering
\small
\vspace{-1em}
    \caption{The effect of granularity in visual observation descriptions.}
    \label{tab:granularity}
    \resizebox{1\linewidth}{!}
    {
    \begin{tabular}{lcccc>{\columncolor{Gray}}c>{\columncolor{Gray}}c}
    \toprule
    \multicolumn{1}{c}{Granularity} & \multicolumn{1}{c}{\#} &
    \multicolumn{1}{c}{TL} & \multicolumn{1}{c}{NE$\downarrow$} & \multicolumn{1}{c}{OSR$\uparrow$} & \multicolumn{1}{c}{SR$\uparrow$} & \multicolumn{1}{c}{SPL$\uparrow$} \\
    \midrule
    FoV@60, 12 views & 1
    & 12.38 & 9.07 & 14.35 & 10.19 & 6.52  \\
    FoV@30, 36 views & 2 
    & 12.67 & 8.92 & 15.28 & 13.89 & 9.12  \\
    FoV@45, 24 views & 3 
    & 12.18 & 8.02 & 26.39 & \color{blue}16.67 & \color{blue}13.00 \\
    \bottomrule
    \end{tabular}
    }
\vspace{-1em}
\end{wraptable}

The Field of View (FoV) of an image critically influences BILP-2's captioning ability, with an overly large FoV leading to generalized room descriptions and an extremely small FoV hindering object recognition due to limited content. As shown in table \ref{tab:granularity}, we investigate 3 granularity of visual representation from a viewpoint. Specifically, variant $\#1$ utilizes an image with 60 FoV, turn heading angle 30 degrees clock-wise to obtain 12 views from a viewpoint, while variant $\#2$ and $\#3$ utilize an image with 30, 45 FoV, turn elevation angle 30 degrees from top to down, and turn heading angle 30, 45 degrees clockwise to form 36 views, 24 views respectively. From the results, we found that using FoV 45 with 24 views for a viewpoint generates the most suitable natural language description for navigation from the BILP-2 model. Using description under such granularity surpasses variant $\#1$ and $\#2$ by $6.48\%$ and $2.78\%$ respectively.

\vspace{-0.5em}
\paragraph{Effect of semantic scene understanding and depth estimation.}
\definecolor{Gray}{gray}{0.9}
\begin{wraptable}{r}{0.55\linewidth}
\vspace{-1em}
    \caption{The effect of semantic scene understanding and depth estimation.}
    \label{tab:obj}
    \centering
    \resizebox{1\linewidth}{!}
    {
    \begin{tabular}{lcccc>{\columncolor{Gray}}c>{\columncolor{Gray}}c}
    \toprule
    \multicolumn{1}{c}{Agent Observation} & \multicolumn{1}{c}{\#} &
    \multicolumn{1}{c}{TL} & \multicolumn{1}{c}{NE$\downarrow$} & \multicolumn{1}{c}{OSR$\uparrow$} & \multicolumn{1}{c}{SR$\uparrow$} & \multicolumn{1}{c}{SPL$\uparrow$} \\
    \midrule
    Baseline & 1
    & 16.11 & 9.83 & 15.28 & 11.11 & 6.92  \\
    Baseline + Obj & 2 
    & 11.07 & 8.88 & 23.34 & 15.97 & 11.71  \\
    Baseline + Obj + Dis & 3 
    & 12.18 & 8.02 & 26.39 & \color{blue}16.67 & \color{blue}13.00 \\
    \bottomrule
    \end{tabular}
    }
    \vspace{-0.5em}
\end{wraptable}

In addition to the granularity of natural language description of the environment, NavGPT also collaborates with other visual foundation models like object detectors and depth estimators to enhance the perception of the current environment. We investigate the effectiveness of adding the object information and the relative distance between the agent and the detected objects. We constructed a baseline method based on the caption results from BILP-2 and powered by GPT-3.5. As shown in table \ref{tab:obj}, by adding object information, the SR increase by $4.86\%$ compared with the baseline, for the additional object information emphasizes the salient object in the scenes. Moreover, we observed a phenomenon in that agents failed to reach the destination because they do not know how close they are to the destination. Once the target viewpoint is visible in sight, they tend to stop immediately. Therefore by adding depth information, the agent has a better understanding of the current position and further rise the SR by $0.7\%$ and SPL by 1.29.

\section{Conclusion}

In this work, we explore the potential of utilizing LLMs in embodied navigation tasks. We present NavGPT, an autonomous LLM system specifically engineered for language-guided navigation, possessing the ability to process multi-modal inputs and unrestricted language guidance, engage with open-world environments, and maintain the navigation history. Limited by the quality of language description of visual scenes and the tracking abilities of objects, NavGPT's zero-shot performance on VLN is still not compatible with trained methods. However, the reasoning trace of GPT-4 illuminates the latent potential of LLMs in embodied navigation planning. Interaction of LLMs with downstream specialized models or the development of multi-modal LLMs for navigation, heralding the future of versatile VLN agents.



{\small
\bibliographystyle{abbrv}
\bibliography{reference}
}

\appendix

\newpage
\section*{Supplementary Material for NavGPT}
Section~\ref{sec:implementation} provides additional details for NavGPT, including each component's prompt and examples of observation descriptions. The experimental setup for prompting GPT-4 to generate instructions and draw top-down trajectory is described in Section~\ref{sec:GPT-4}. Section~\ref{sec:fail} illustrates the limitation of NavGPT with some failure cases. Finally, Section~\ref{sec:broader} discusses the broader impacts of our work.

\section{Implementation Details}
\label{sec:implementation}

\subsection{Convert Visual Perception to Language Description (\S 3.2\protect\footnote{Refer to section \ref{sec:VFM} in main paper.})}
\begin{figure}[h]
	\centering
    \includegraphics[width=.99\linewidth]{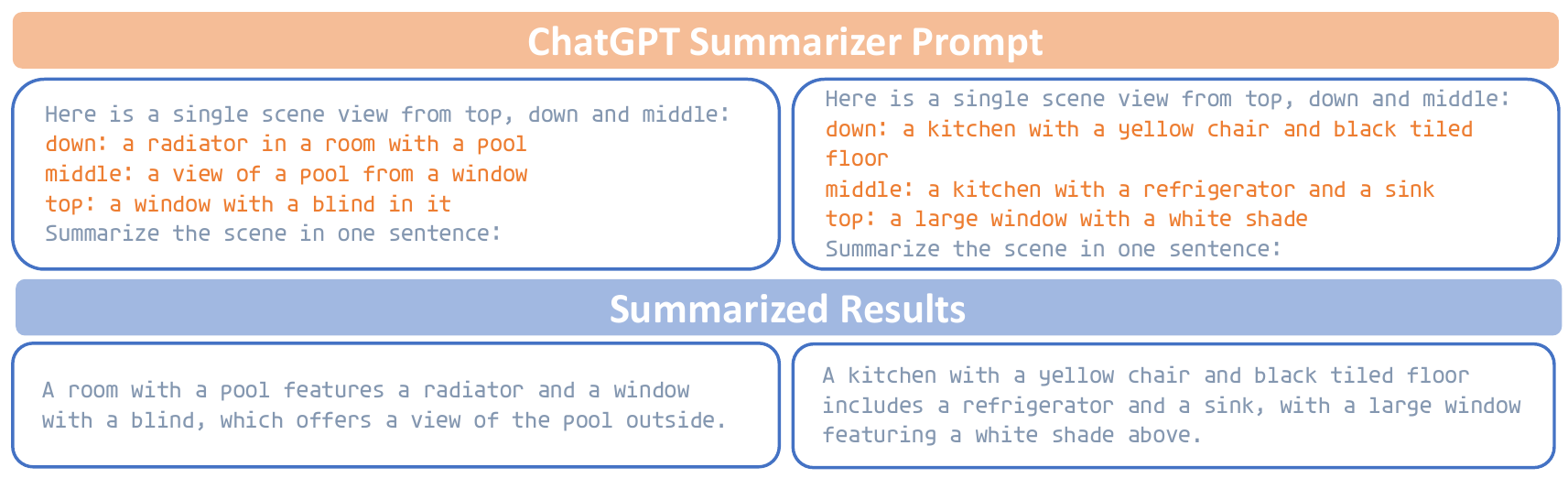}
	\caption{\small The prompt for GPT-3.5 summarizer and the summarized results. The original descriptions from BLIP-2 are in orange.}
	\label{fig:Bilp}
\vspace{-1em}
\end{figure}

For each viewpoint, given a heading direction $\theta$, we use elevation angles $-30^{\circ}$, $0^{\circ}$ and $30^{\circ}$ to capture three egocentric images from down, middle and top to form the observation for this direction. The field of view of each image is $45^{\circ}$, so there is an overlapping of $15^{\circ}$ of the there images in the same direction. The visual perception process for each direction includes two steps, including using BILP-2~\cite{li2023blip} to caption the three images, then, summarizing the descriptions by the GPT-3.5 summarizer.
\paragraph{BILP-2 Prompt}
We tried various ways to prompt the BILP-2 model given the images from a viewpoint. Such as no prompt, prompting it with "\textit{Detailly describe the scene.}" or "\textit{This is a scene of}". Ultimately, we selected "\textit{This is a scene of}" as the preferred prompt for BILP-2 to generate descriptions for each image. Utilizing no prompt can lead to inconsistent description lengths, occasionally rendering the description excessively brief. When applying the prompt "\textit{Detailly describe the scene.}", the resulting description primarily centers on the room type, neglecting object details. In contrast, our chosen prompt yields language descriptions that are highly pertinent to indoor scenes and emphasize object depictions. The examples of the caption results are shown in figure \ref{fig:Bilp}.

\paragraph{GPT-3.5 Summarizer Prompt}
Descriptions from BILP-2 could have a substantial amount of redundancy because the same object could show up in the three images simultaneously. We adopt a GPT-3.5 summarizer to summarize them into one sentence following the template: "\textit{Here is a single scene view from top, down and middle:\textbackslash n\{description\}\textbackslash nSummarize the scene in one sentence:}", where the "\textit{\{description\}}" is replaced with the generated text of top, middle and down images from BILP-2, shown in figure \ref{fig:Bilp}.

\begin{figure}[h]
	\centering
    \includegraphics[width=.99\linewidth]{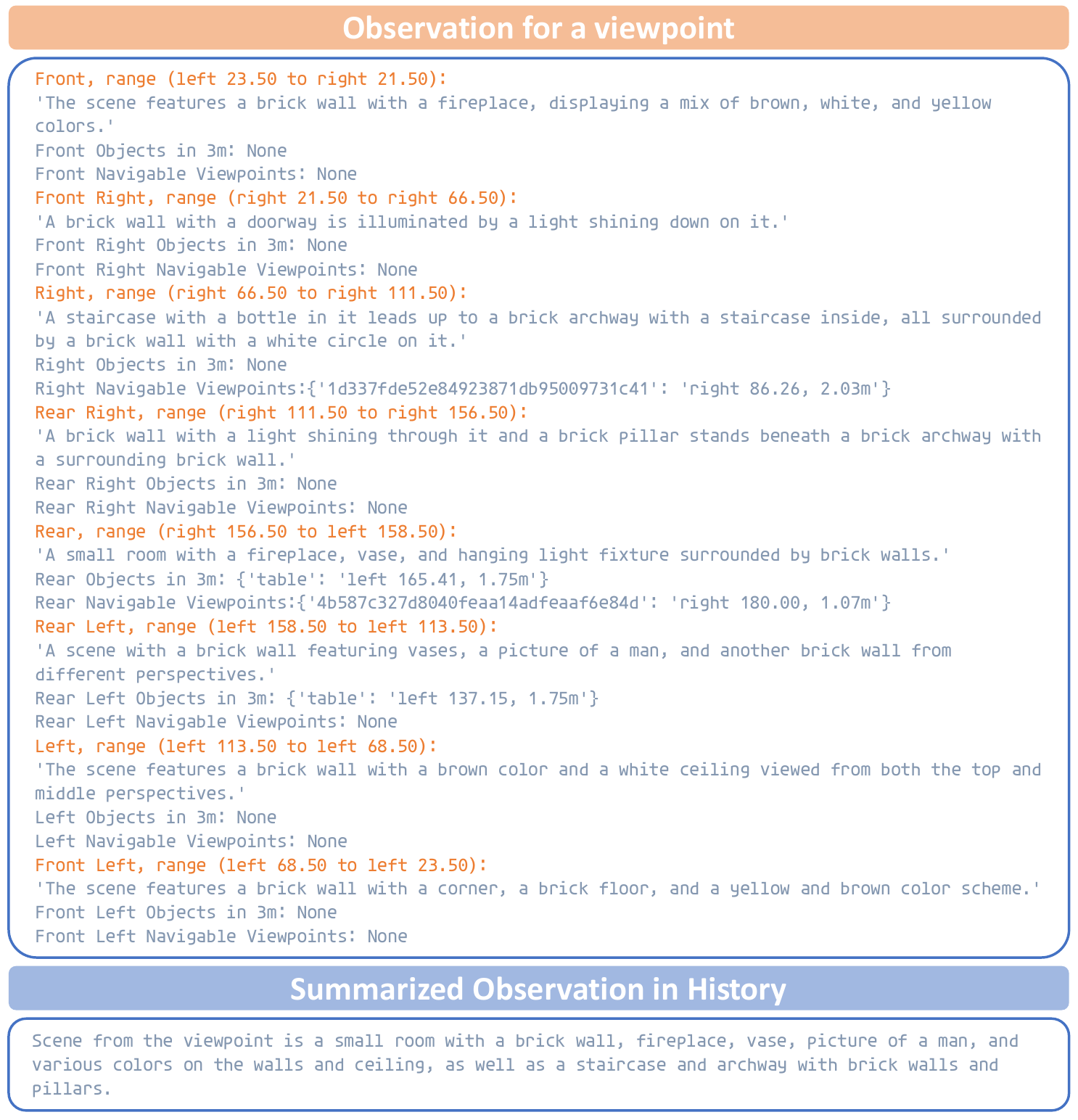}
	\caption{\small The language description of observation for a single viewpoint.}
	\label{fig:obseravtion_vp}
\vspace{-1em}
\end{figure}

\newpage
\paragraph{Observation description examples}
Given the summarized description of each direction, along with the objects detected from the object detector, the agent interacts with the Matterport Simulator~\cite{anderson2018vision} to extract the depth and the navigable viewpoints information. The prompt manager will take the current heading of the agent as the "front" direction, and calculate the relative angle between the agent's current heading and the detected objects as well as the navigable viewpoints, concatenating the descriptions from each direction clockwise. The overall observation for a single viewpoint is shown in figure \ref{fig:obseravtion_vp}.

\subsection{NavGPT Prompt (\S 3.4)}

\begin{figure}[h]
	\centering
    \includegraphics[width=.99\linewidth]{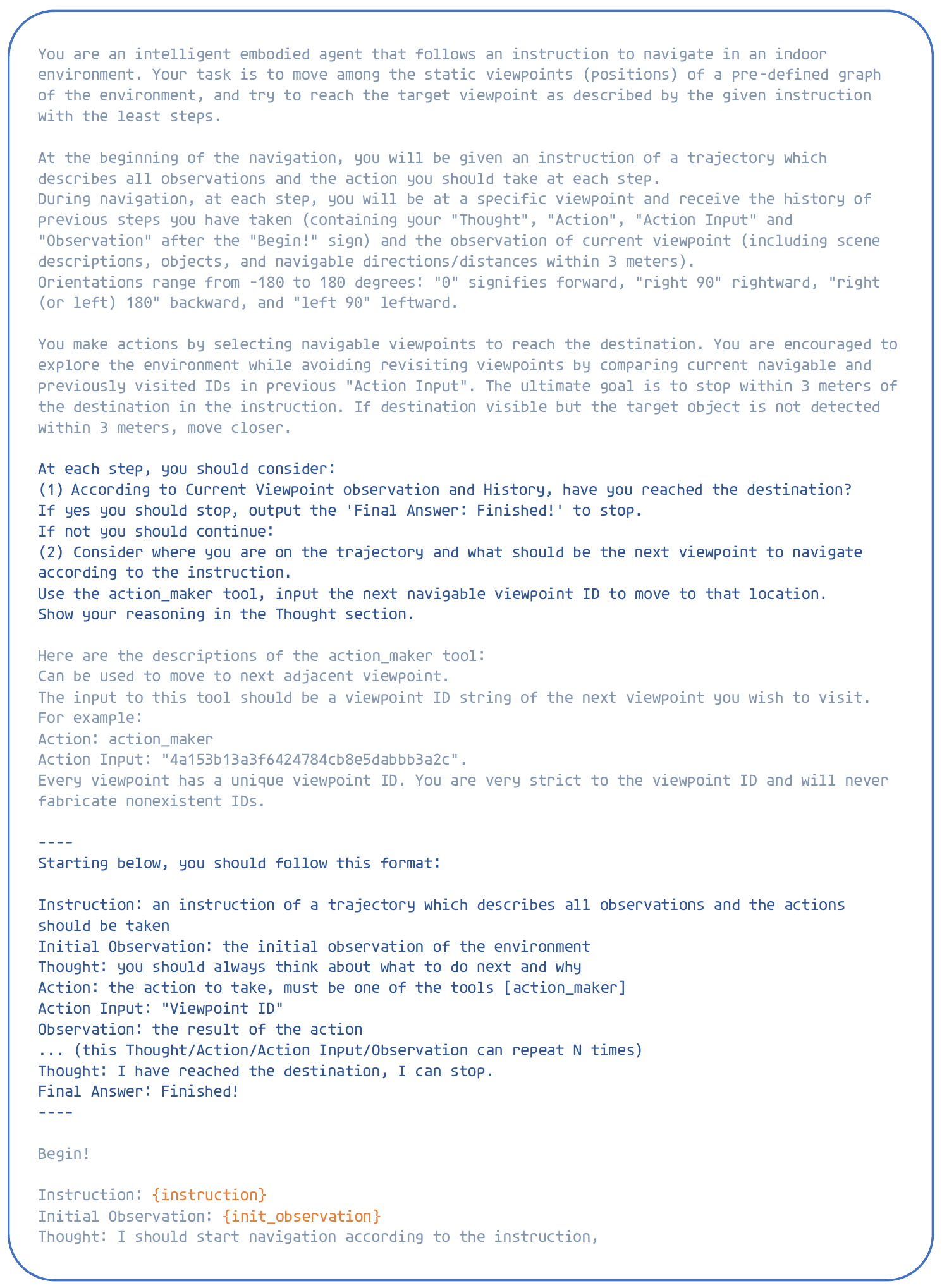}
	\caption{\small Navigation system principle prompt.}
	\label{fig:principle}
\vspace{-1em}
\end{figure}

\paragraph{Navigation System Principle}
The Navigation system principle for NavGPT is shown in figure \ref{fig:principle}, it defines the VLN task and the basic reasoning format and rules for NavGPT at each navigation step.

The "\textit{\{instruction\}}" and "\textit{\{init\_observation\}}" in figure \ref{fig:principle} will be replaced with the specific instruction and the language description of the starting point respectively.

\paragraph{History with Summarizer}
For history during navigation, directly using the description shown in figure \ref{fig:obseravtion_vp} will be too long for LLMs to accept. We adopt a GPT-3.5 summarizer to summarize the observation into a sentence to replace the $\mathcal{O}_{<t+1}$ in $\mathcal{H}_{<t+1}$. The prompt template we used is: "\textit{Given the description of a viewpoint. Summarize the scene from the viewpoint in one concise sentence.\textbackslash nDescription:\textbackslash n\{description\}\textbackslash n Summarization: The scene from the viewpoint is a}". where the "\textit{\{description\}}" is replaced with the textual description of the viewpoint. The summarized observation for viewpoints is shown in figure \ref{fig:obseravtion_vp}. The observation in history $\mathcal{O}_{<t+1}$ is the summarized observation of viewpoint, while the current viewpoint observation $\mathcal{O}_{t+1}$ is the full description.

\section{Prompting GPT-4 to Generate Instruction and Draw Top-down Trajectory (\S 4.1)}
\label{sec:GPT-4}
\begin{figure}[h]
	\centering
    \includegraphics[width=.99\linewidth]{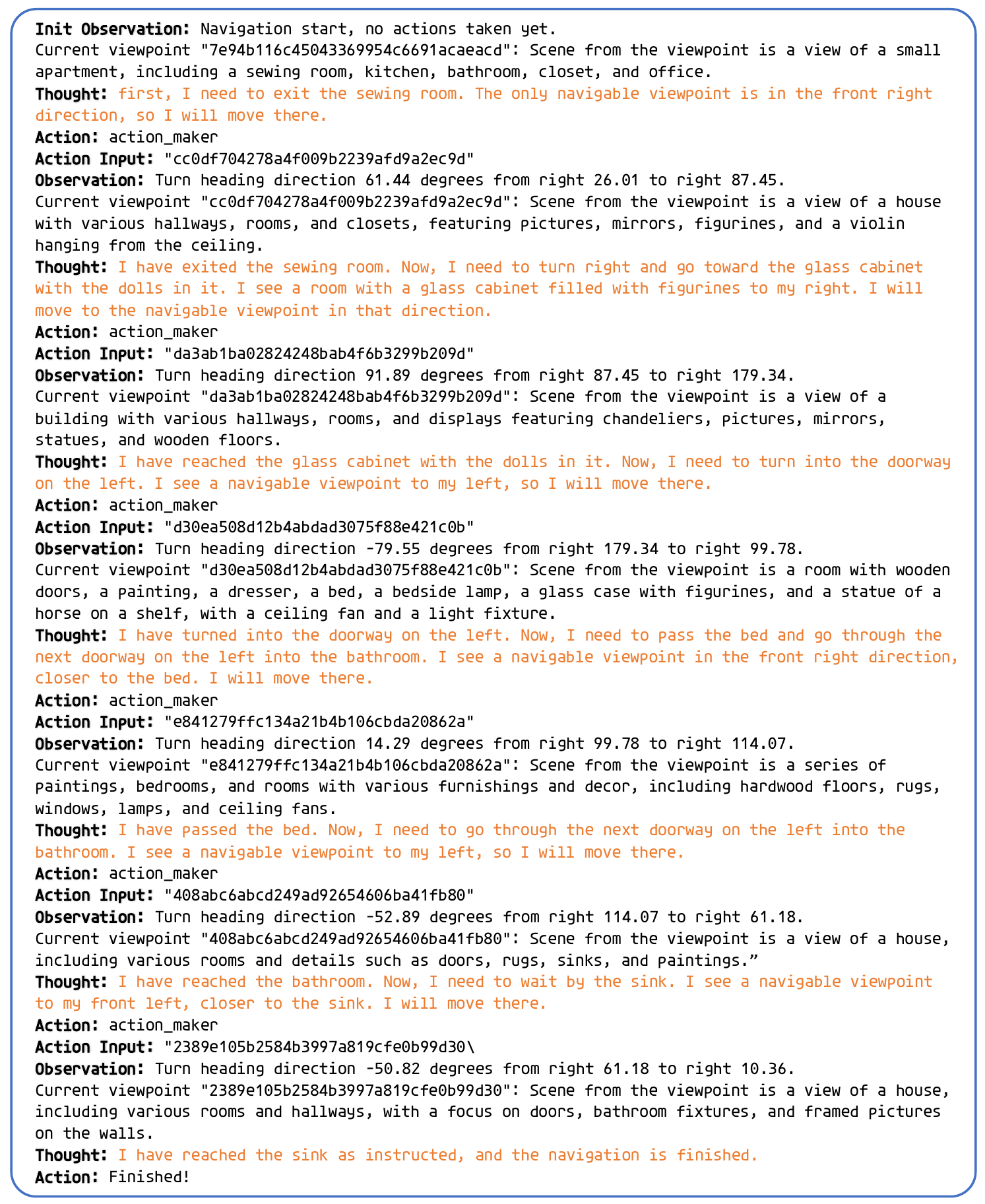}
	\caption{\small The history $\mathcal{H}_{<t+1}$ of NavGPT.}
	\label{fig:traj}
\vspace{-1em}
\end{figure}

\begin{figure}[h]
	\centering
    \includegraphics[width=.99\linewidth]{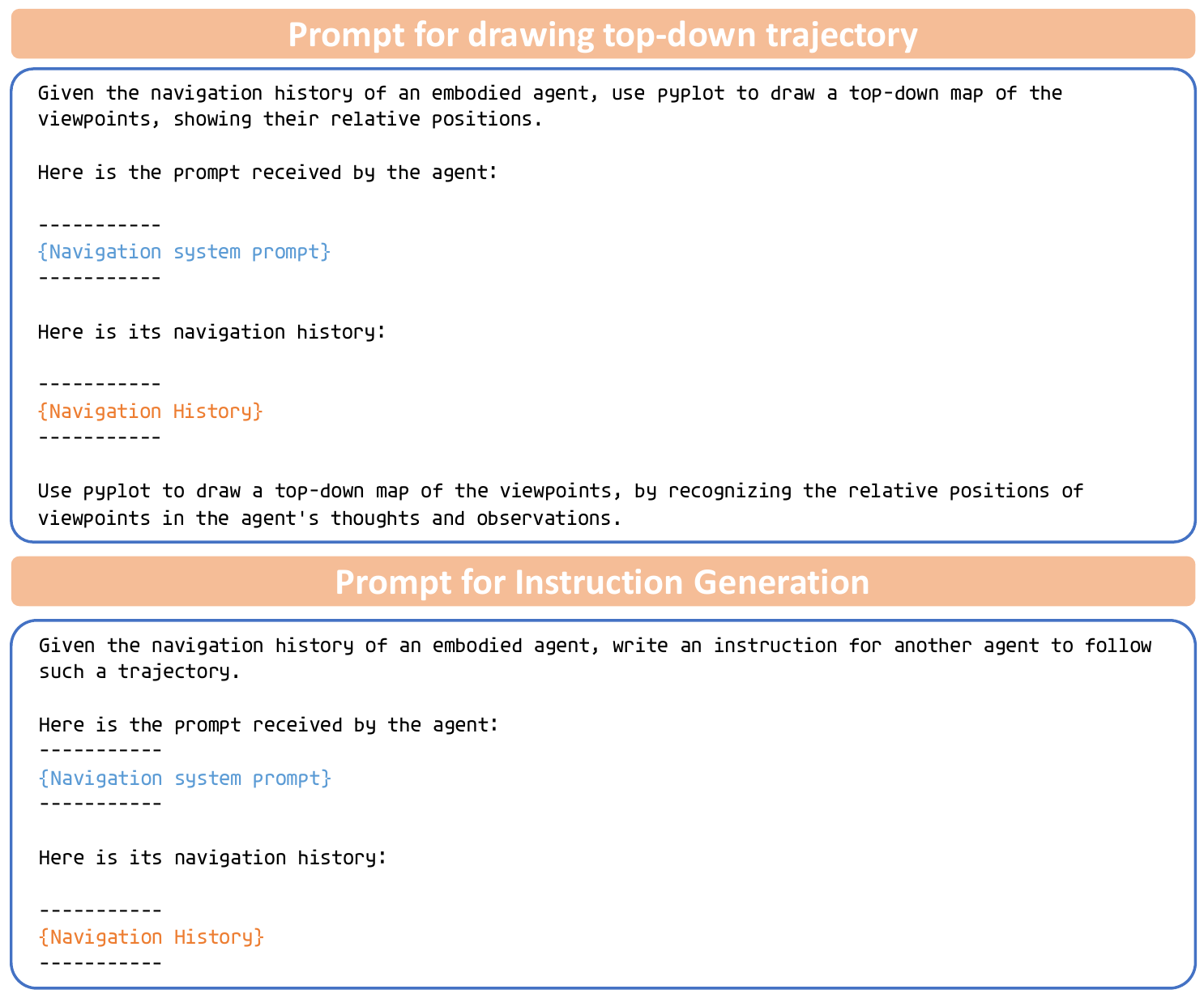}
	\caption{\small Prompts for GPT-4 to generate instruction and draw top-down trajectory.}
	\label{fig:GPT-4 prompt}
\vspace{-1em}
\end{figure}

We choose a case where NavGPT successfully followed the ground truth trajectory, the history of NavGPT is shown in figure \ref{fig:traj}. The prompts for GPT-4 are shown in figure \ref{fig:GPT-4 prompt}.

\newpage

\begin{figure}[h]
	\centering
    \includegraphics[width=.94\linewidth]{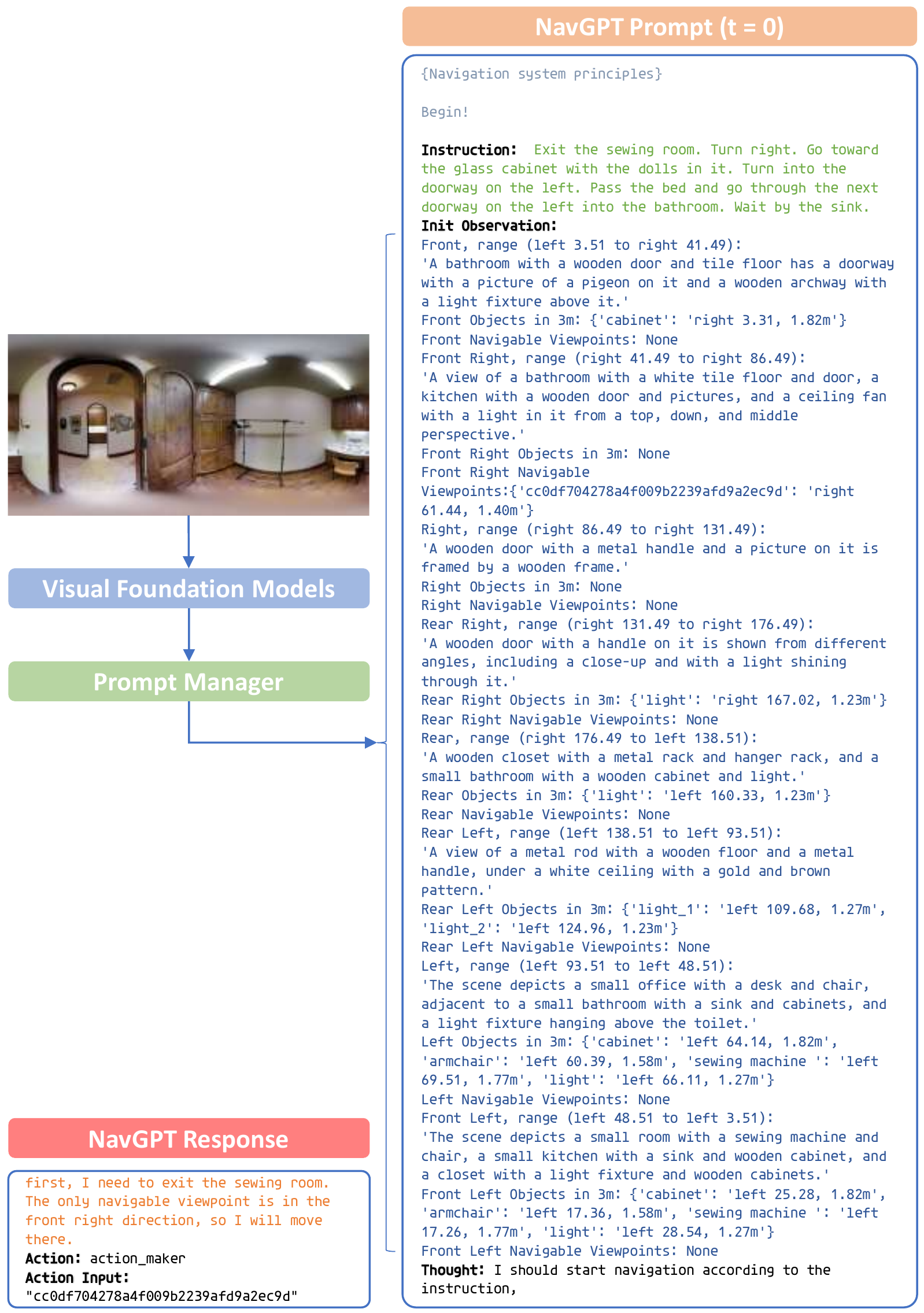}
	\caption{\small The prompt and response of NavGPT at step 0. All the text in NavGPT's response is generated by GPT-4. The "\textit{\{Navigation system principles\}}" is shown in figure \ref{fig:principle}.}
	\label{fig:c1}
\vspace{-2em}
\end{figure}

\begin{figure}[h]
	\centering
    \includegraphics[width=.98\linewidth]{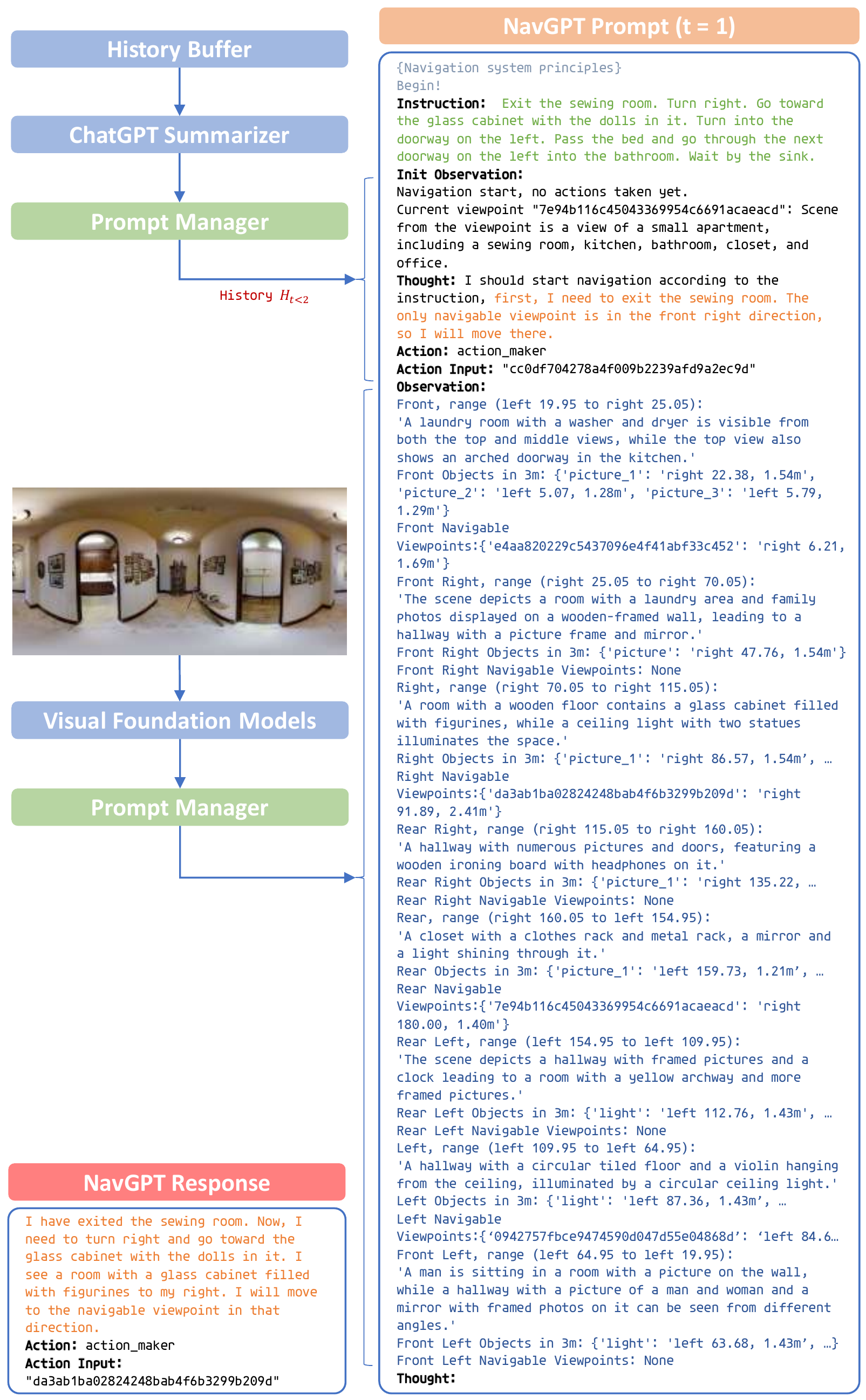}
	\caption{\small The prompt and response of NavGPT at step 1. All the text in NavGPT's response is generated by GPT-4. The "\textit{\{Navigation system principles\}}" is shown in figure \ref{fig:principle}.}
	\label{fig:c2}
\vspace{-1em}
\end{figure}

\begin{figure}[h]
	\centering
    \includegraphics[width=.97\linewidth]{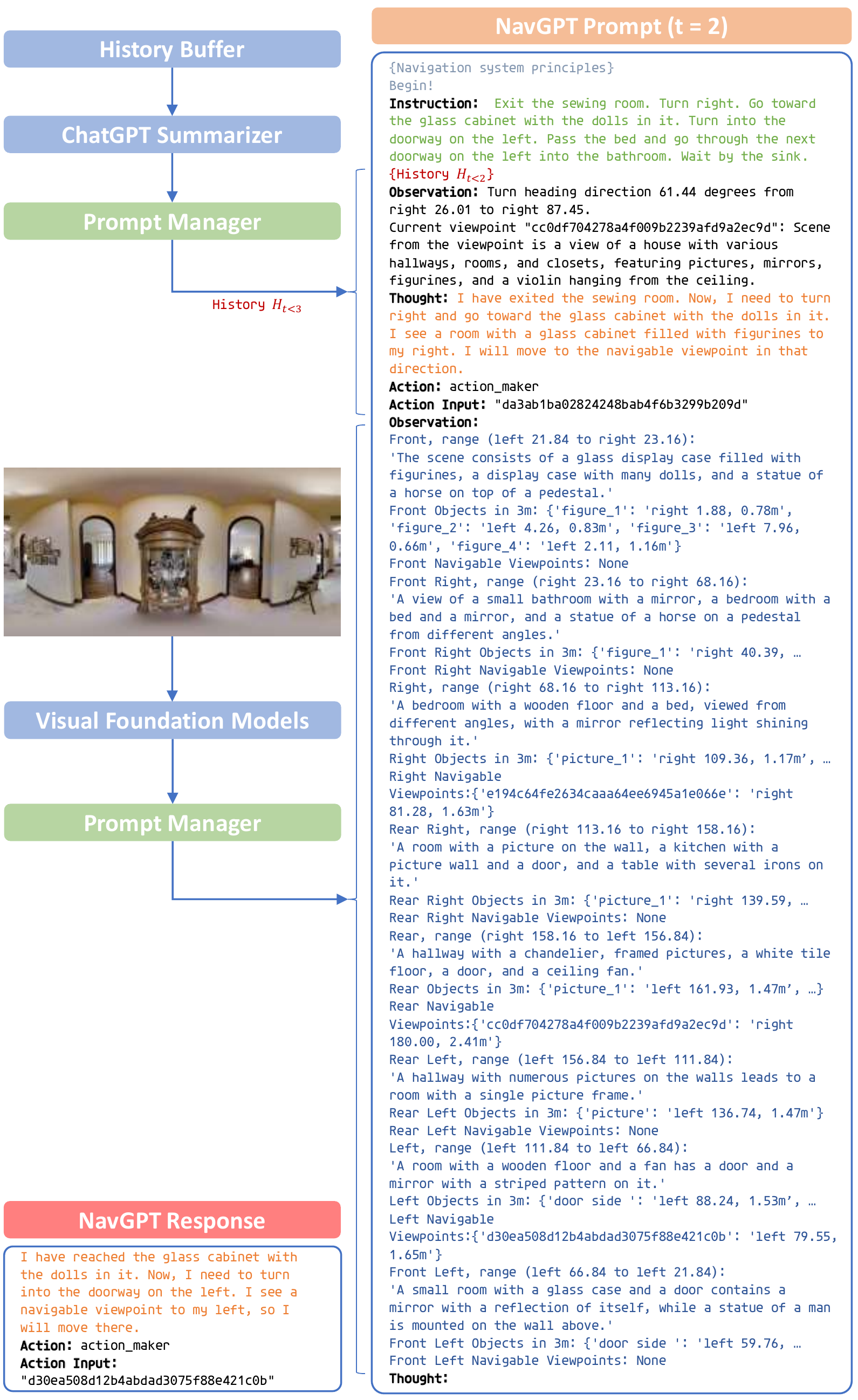}
	\caption{\small The prompt and response of NavGPT at step 2. All the text in NavGPT's response is generated by GPT-4. The "\textit{\{Navigation system principles\}}" is shown in figure \ref{fig:principle}, the "\textit{\{History $\mathcal{H}_{<2}$\}}" is shown in figure \ref{fig:c2}.}
	\label{fig:c3}
\vspace{-1em}
\end{figure}

\begin{figure}[h]
	\centering
    \includegraphics[width=.97\linewidth]{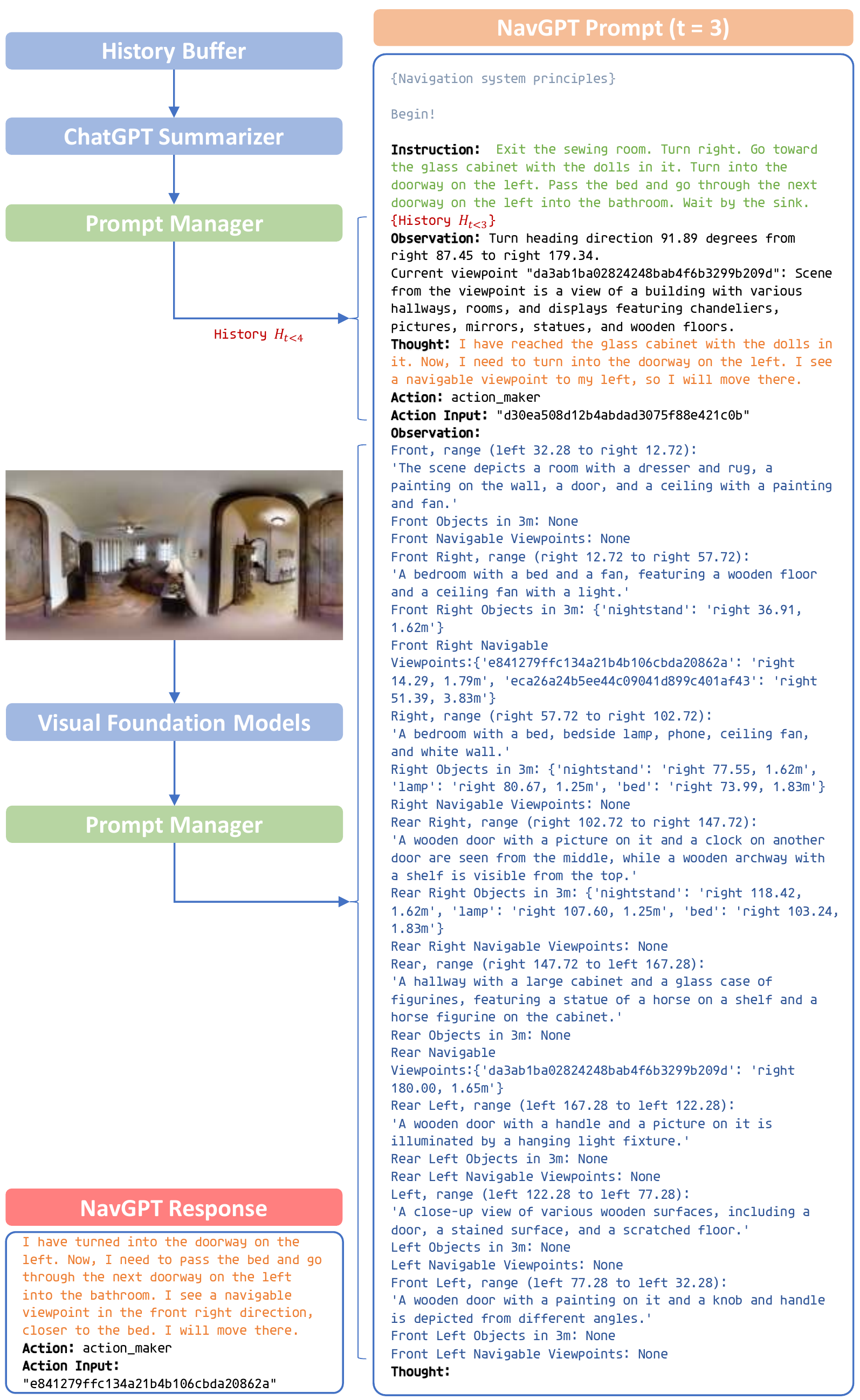}
	\caption{\small The prompt and response of NavGPT at step 3. All the text in NavGPT's response is generated by GPT-4. The "\textit{\{Navigation system principles\}}" is shown in figure \ref{fig:principle}, the "\textit{\{History $\mathcal{H}_{<3}$\}}" is shown in figure \ref{fig:c3}.}
	\label{fig:c4}
\vspace{-1em}
\end{figure}

\begin{figure}[h]
	\centering
    \includegraphics[width=.97\linewidth]{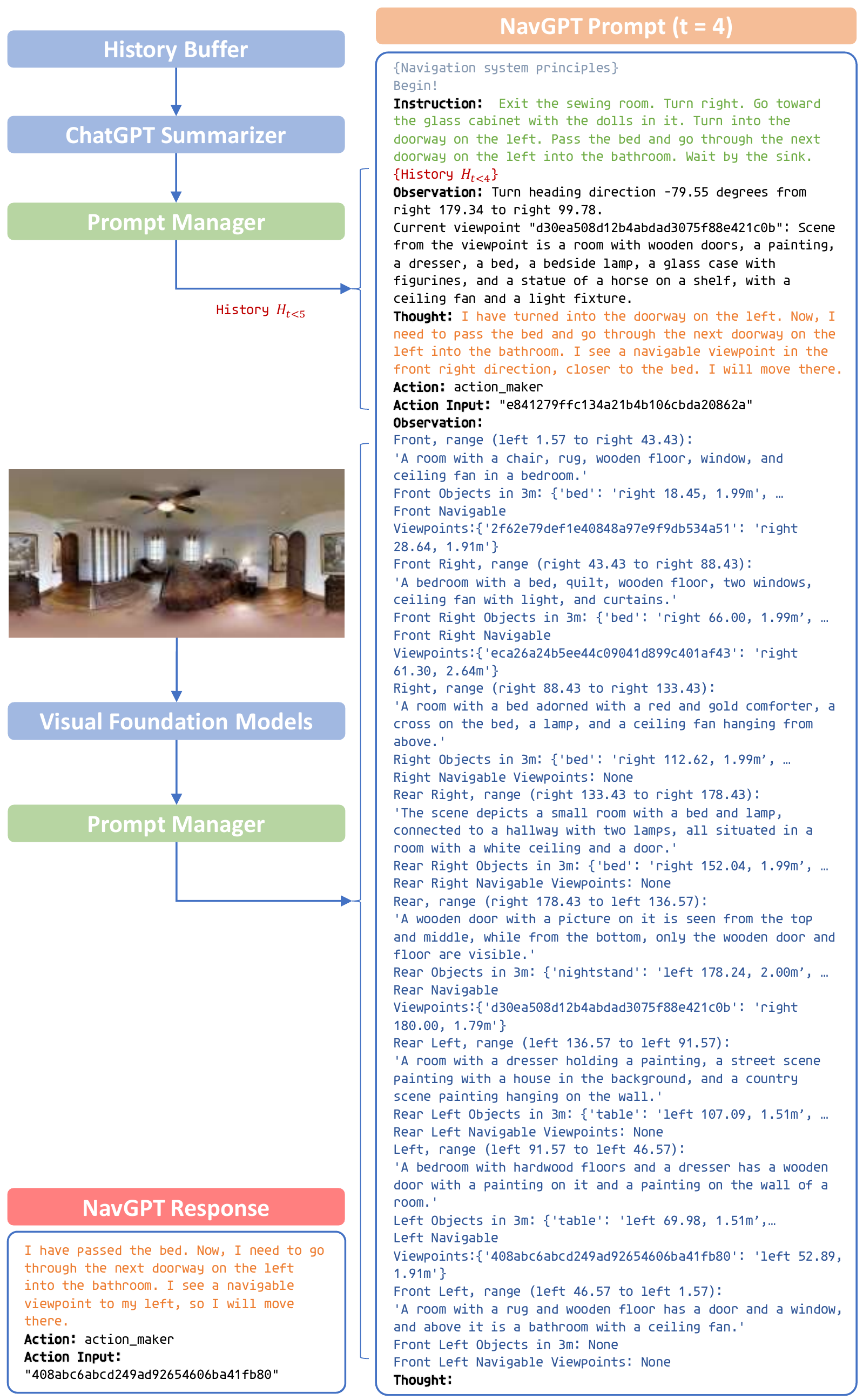}
	\caption{\small The prompt and response of NavGPT at step 4. All the text in NavGPT's response is generated by GPT-4. The "\textit{\{Navigation system principles\}}" is shown in figure \ref{fig:principle}, the "\textit{\{History $\mathcal{H}_{<4}$\}}" is shown in figure \ref{fig:c4}.}
	\label{fig:c5}
\vspace{-1em}
\end{figure}

\begin{figure}[h]
	\centering
    \includegraphics[width=.97\linewidth]{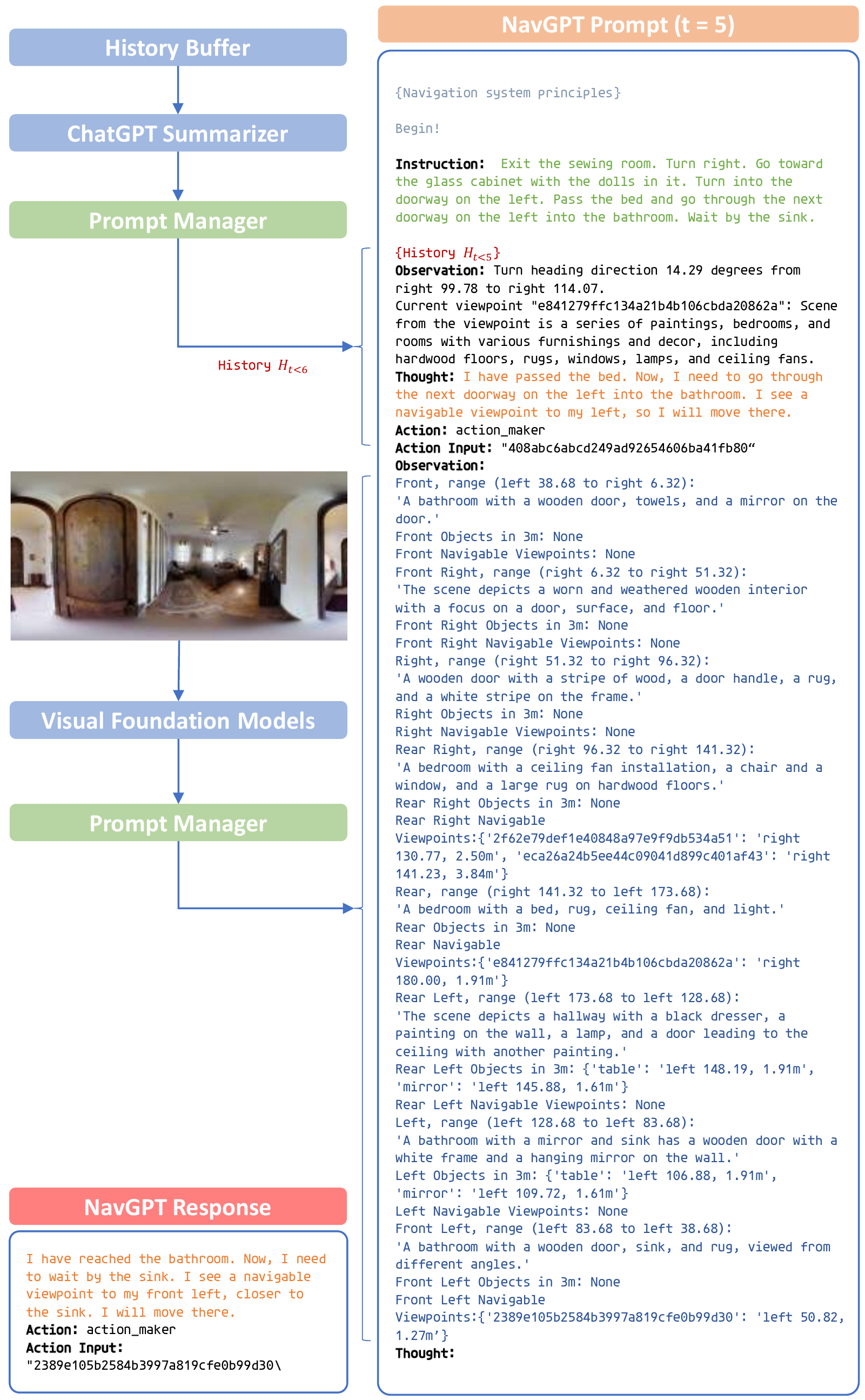}
	\caption{\small The prompt and response of NavGPT at step 5. All the text in NavGPT's response is generated by GPT-4. The "\textit{\{Navigation system principles\}}" is shown in figure \ref{fig:principle}, the "\textit{\{History $\mathcal{H}_{<5}$\}}" is shown in figure \ref{fig:c5}.}
	\label{fig:c6}
\vspace{-1em}
\end{figure}

\begin{figure}[h]
	\centering
    \includegraphics[width=.97\linewidth]{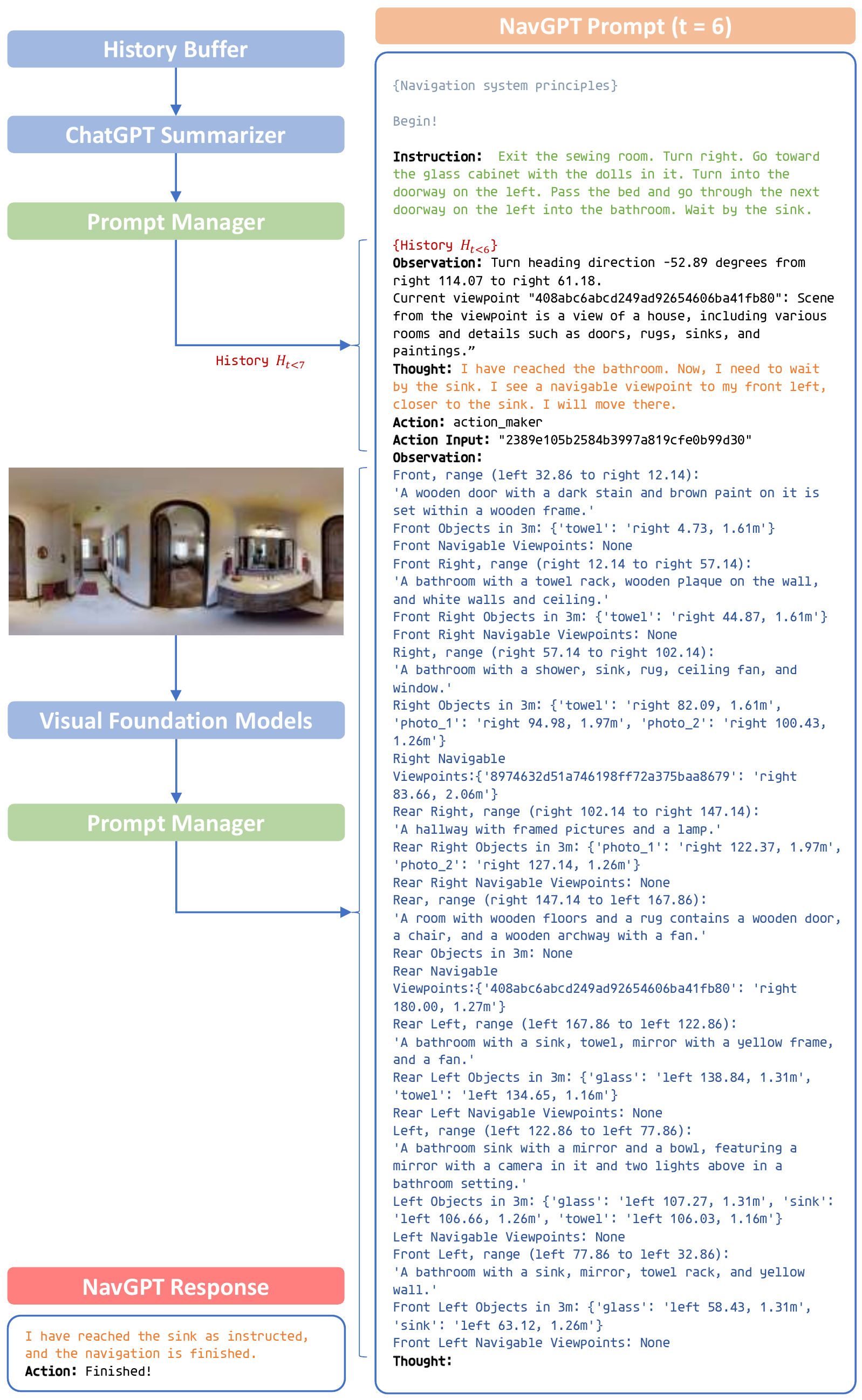}
	\caption{\small The prompt and response of NavGPT at step 6. All the text in NavGPT's response is generated by GPT-4. The "\textit{\{Navigation system principles\}}" is shown in figure \ref{fig:principle}, the "\textit{\{History $\mathcal{H}_{<6}$\}}" is shown in figure \ref{fig:c6}.}
	\label{fig:c7}
\vspace{-1em}
\end{figure}

We further show the detail of prompts and NavGPT's response at each step in this example in figure~\ref{fig:c1}--\ref{fig:c7}. The beginning of each prompt, the "\textit{\{Navigation system principles\}}" is replaced with the Navigation system principle prompt shown in figure \ref{fig:principle}.

\section{NavGPT Failure cases (\S 4.2)}
\label{sec:fail}
In this section, we show some failure cases of NavGPT to illustrate the limitation of our method, specifically pertaining to the information degradation in the linguistic representation of visual scenes and the object tracking abilities.

\begin{figure}[h]
	\centering
    \includegraphics[width=.99\linewidth]{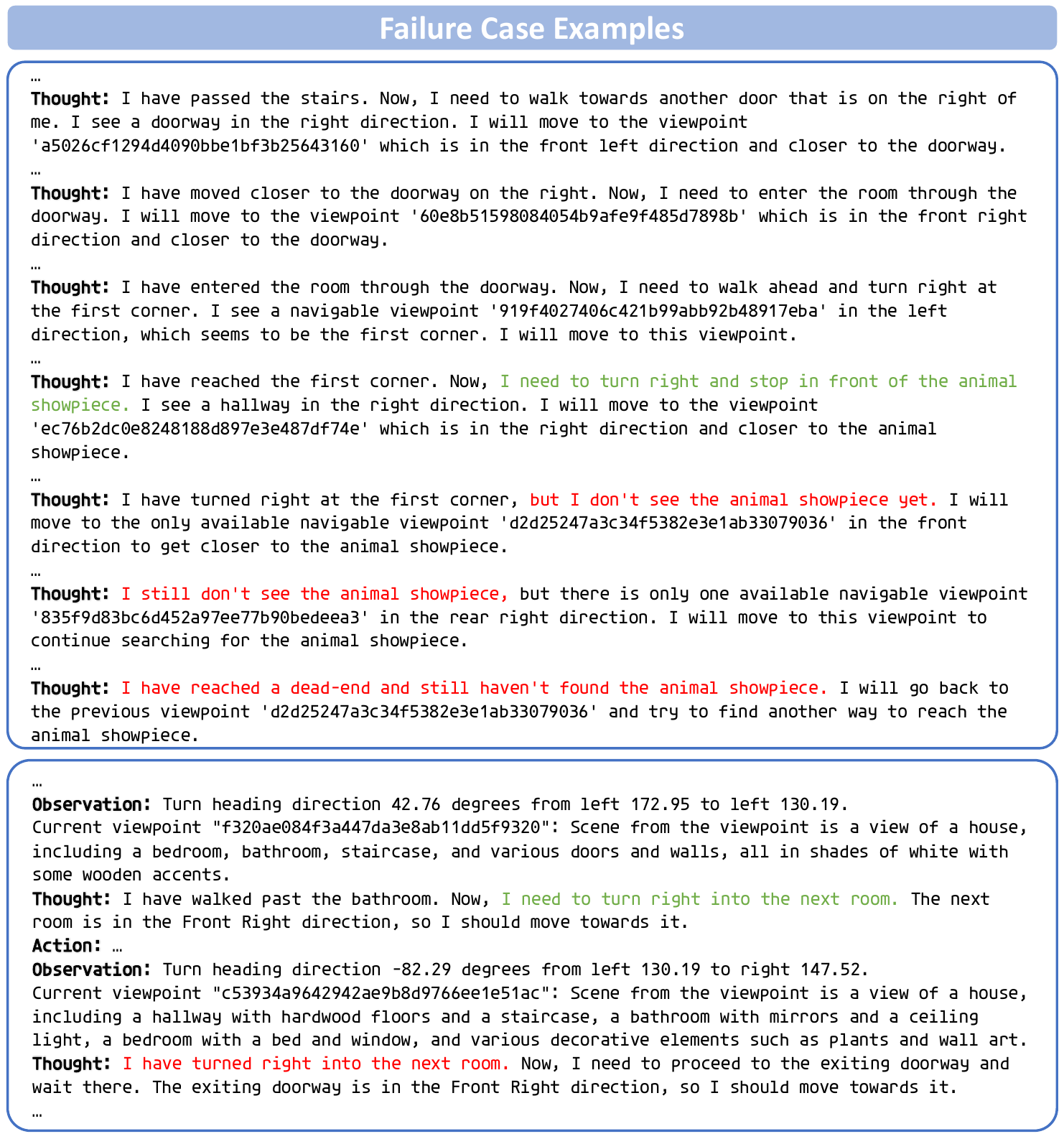}
	\caption{\small Failure cases in NavGPT.}
	\label{fig:fail}
\vspace{-1em}
\end{figure}

As depicted in Figure \ref{fig:fail}, the top example demonstrates that if the target object outlined in the instruction is missing from the observation description, NavGPT is necessitated to explore the environment. Ideally, all target objects should be discernible in accordance with the instruction. Nonetheless, the inherent granularity of language description compared to visual perception invariably results in information loss. In the lower example, the agent is required to "\textit{turn right into the next room}". However, since the historical viewpoint observations are summarized into a sentence, some details have been omitted. NavGPT cannot compare with the comprehensive observation descriptions between viewpoints to construct a continuous understanding of the environment. The lacking a reference substance between viewpoints prevents NavGPT from determining whether it has completed the previous action. In the lower example, the agent has not entered the room yet while it erroneously believed that "\textit{I have turned right into the next room}".

A potential way to solve the problem is to generate the language descriptions dynamically, similar to the ChatCaptioner\cite{zhu2023chatgpt}. Unlike the current implementation, where visual information is compromised during translation into static language descriptions, through the interaction between LLMs~\cite{zhu2023chatgpt}, the agent can extract essential information from the environment during navigation.

\section{Broader Impact}
\label{sec:broader}
Our work is the initial attempt to use GPT models toward building versatile VLN agents. We believe the reasoning capability of LLMs is the foundation for generalizable embodied navigation agents. NavGPT reveals the reasoning trace of LLMs during navigation, making the process explicit and explainable. For safety and ethical concerns, at the current stage, all the experiments are done on the open-source Vision-and-Language Navigation dataset in a simulated environment, which ensures the controllability of the agent's behaviors. At the same time, for exploration of the practical implementation of this technology in the future, the robustness of the performance of generative models cannot yet be guaranteed. Further research is required including how to prompt LLMs to increase the accuracy and precision in planning and sequential action predictions, which is the key consideration for safety issues in real-world deployment.

\end{document}